\documentclass[compsoc,conference,a4paper,10pt,times]{IEEEtran}
\IEEEoverridecommandlockouts
\usepackage{cite}
\usepackage{amsmath,amssymb,amsfonts}
\usepackage{algorithmic}
\usepackage{graphicx}
\usepackage{textcomp}
\usepackage{bmpsize}
\usepackage{xcolor}
\usepackage{lipsum}
\usepackage[colorlinks=true,urlcolor=black]{hyperref}
\newcommand{\mainchange}[0]{black}
\def\BibTeX{{\rm B\kern-.05em{\sc i\kern-.025em b}\kern-.08em
    T\kern-.1667em\lower.7ex\hbox{E}\kern-.125emX}}

\usepackage{amsmath}
\usepackage{amssymb}
\usepackage{mathtools}
\usepackage{amsthm}
\usepackage{svg}
\usepackage{booktabs}
\usepackage{multirow}

\usepackage{enumitem}

\usepackage{pifont} %
\usepackage[margin=1in]{geometry}

\usepackage[capitalize,noabbrev]{cleveref}

\theoremstyle{plain}

\theoremstyle{definition}

\theoremstyle{remark}

\graphicspath{{figures/}}

\usepackage[acronym]{glossaries}

\newcommand{\nmethods}{$18 \text{ }$}
\newcommand{\ndatasets}{$5 \text{ }$}
\newcommand{\narchitectures}{$2 \text{ }$}

\newacronym{unl_1_fcs}{FCS}{Forget-Contrast-Strengthen}
\newacronym{unl_2_msg}{MSG}{Masked-Small-Gradients}
\newacronym{unl_3_cfw}{CFW}{Confuse-Finetune-Weaken}
\newacronym{unl_4_prmq}{PRMQ}{Prune-Reinitialize-Match-Quantize}
\newacronym{unl_5_ct}{CT}{Convolution-Transpose}
\newacronym{unl_6_kde}{KDE}{Knowledge-Distillation-Entropy}
\newacronym{unl_7_rni}{RNI}{Repeated-Noise-Injection}
\newacronym{unl_bt}{BT}{Bad-Teacher}
\newacronym{unl_ft}{FT}{Fine-Tune}
\newacronym{unl_srl}{SRL}{Successive Random Labels}
\newacronym{unl_ga}{GA}{Gradient Ascent}
\newacronym{unl_scrub}{SCRUB}{SCalable Remembering and Unlearning unBound}
\newacronym{unl_euk}{EU-K}{Exact Unlearning - K}
\newacronym{unl_cfk}{CF-K}{Catastrophic Forgetting - K}
\newacronym{unl_iu}{IU}{Influence Unlearning}
\newacronym{unl_fisher}{FF}{Fisher Forgetting}
\newacronym{unl_salun}{SaLUN}{Saliency Unlearning}
\newacronym{unl_ng+}{NG+}{Negative Gradient Plus}

\newacronym{munl}{MU}{Machine Unlearning}
\newacronym{dl}{DL}{Deep Learning}
\newacronym{ml}{ML}{Machine Learning}
\newacronym{mia}{MIA}{Membership Inference Attack}
\newacronym{dnn}{DNN}{Deep Neural Network}
\newacronym{kld}{KLD}{Kullback-Leibler Divergence}
\newacronym{jsd}{JSD}{Jensen-Shannon Divergence}
\newacronym{cnn}{CNN}{Convolutional Neural Network}
\newacronym{cosine}{CS}{Cosine Similarity}
\newacronym{ranking_method}{MBC}{Modified Borda Count}

\renewcommand{\acrshort}[1]{\glsentryshort{#1}}
\renewcommand{\acrfull}[1]{\glsentrylong{#1} (\glsentryshort{#1})}

\begin{document}

\title{Deep Unlearn: Benchmarking Machine Unlearning for Image Classification}

\author{
\IEEEauthorblockN{Xavier F. Cadet}
\IEEEauthorblockA{Imperial College London \\
United Kingdom \\
xavier.cadet17@imperial.ac.uk}
\and
\IEEEauthorblockN{Anastasia Borovykh}
\IEEEauthorblockA{Imperial College London \\
United Kingdom \\
a.borovykh@imperial.ac.uk}
\and
\IEEEauthorblockN{Mohammad Malekzadeh}
\IEEEauthorblockA{Nokia Bell Labs \\
United Kingdom \\
mohammad.malekzadeh@nokia-bell-labs.com}

\and
\IEEEauthorblockN{Sara Ahmadi-Abhari}
\IEEEauthorblockA{Imperial College London \\
United Kingdom \\
s.ahmadi-abhari@imperial.ac.uk}
\and
\IEEEauthorblockN{Hamed Haddadi}
\IEEEauthorblockA{Imperial College London \\
United Kingdom \\
h.haddadi@imperial.ac.uk}
}

\maketitle

\begin{abstract}
Machine unlearning (MU) aims to remove the influence of particular data points from the learnable parameters of a trained machine learning model.
This is a crucial capability in light of data privacy requirements, trustworthiness, and safety in deployed models.
MU is particularly challenging for deep neural networks (DNNs), such as convolutional nets or vision transformers, as such DNNs tend to memorize a notable portion of their training dataset.
Nevertheless, the community lacks a rigorous and multifaceted study that looks into the success of MU methods for DNNs.
In this paper, we investigate 18 state-of-the-art MU methods across various benchmark datasets and models, with each evaluation conducted over 10 different initializations, a comprehensive evaluation involving MU over 100K models.
We show that, with the proper hyperparameters, Masked Small Gradients (MSG) and Convolution Transpose (CT), consistently perform better in terms of model accuracy and run-time efficiency across different models, datasets, and initializations, assessed by population-based membership inference attacks (MIA) and per-sample unlearning likelihood ratio attacks (U-LiRA).
Furthermore, our benchmark highlights the fact that comparing a MU method only with commonly used baselines, such as Gradient Ascent (GA) or Successive Random Relabeling (SRL), is inadequate, 
and we need better baselines like Negative Gradient Plus (NG+) with proper hyperparameter selection.
\end{abstract}


\section{Introduction} \label{section:introduction}

\begin{figure*}
    \centering
    \includegraphics[width=\linewidth]{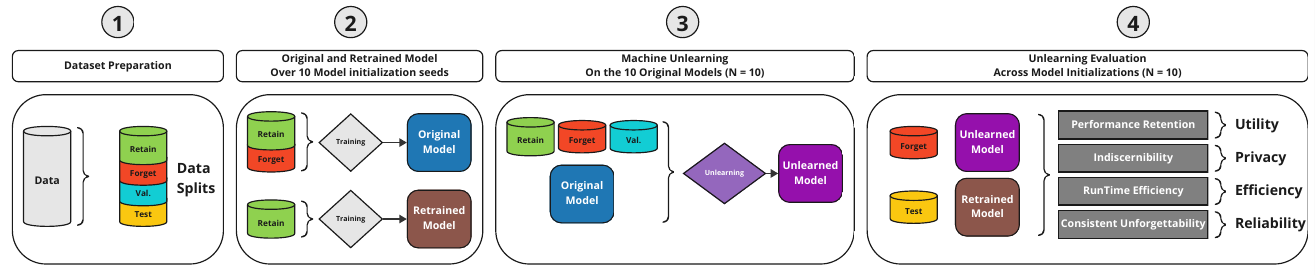}
    \caption{Pipeline on a single dataset. Phase 1: The dataset is split into the data to retain and forget, and evaluation splits. Phase 2: The Original and Retrained models (Reference) are trained across ten shared model initialization seeds. Phase 3: The Machine Unlearning methods are applied to the Original model after three searches with 100 attempts to determine the best hyper-parameters. Phase 4: The model post-unlearning is evaluated with and without the Retrained model across four aspects: Utility, Privacy, Efficiency, and Reliability.}
    \label{figure:flowchart}
\end{figure*}

Machine unlearning aims to remove the influence of a specified subset of training data points from trained models~\cite{cao2015towards}. This process is crucial for enhancing privacy preservation, model safety, and overall model quality. MU helps ensure compliance with the right to be forgotten~\cite{ginart2019making}, removes erroneous data points that negatively impact model performance~\cite{shaikExploringLandscapeMachine2023}, and eliminates biases introduced by parts of the training data~\cite{chenFastModelDeBias2023}.
Deep neural networks present significant challenges for MU due to their computationally intensive training requirements, highly non-convex loss landscapes, and tendency to memorize substantial portions of their training data~\cite{fredriksonModelInversionAttacks2015, carliniSecretSharerEvaluating2019, feldmanWhatNeuralNetworks2020}.

The key open challenges in MU include:
(1) A degradation in model accuracy often accompanies unlearning;
(2) Some MU methods require the model to be trained in specific ways, such as saving checkpoints, tracking accumulated gradients, or training with differential privacy, limiting their applicability to already deployed models;
(3) Assurance of information removal is difficult as there are no reliable metrics to measure it accurately;
(4) There is no consensus on the most effective methods.

A branch of MU known as {\em exact unlearning} aims to guarantee that the specified {\em forget} data have been completely removed from the model.
The most reliable exact MU method is
to {\em retrain} the model from scratch while excluding the forget data.
Another exact MU that offers data removal guarantees is SISA (Sharded, Isolated, Sliced, Aggregated)~\cite{bourtouleMachineUnlearning2021}.
However, exact MU is computationally prohibitive, emphasizing the need for more efficient methods.
The alternative branch is {\em approximate unlearning} that aims to approximate data deletion and is often less precise but more computationally efficient.
Approximate MU lacks theoretical guarantees, necessitating empirical evaluations to determine their effectiveness, reliability, and computational efficiency across various datasets.
Thus, the community needs a proper benchmarking and evaluation of state-of-the-art approximate MU methods.

In this paper, we benchmark \nmethods state-of-the-art approximate MU methods across \ndatasets datasets and \narchitectures DNN architectures commonly used in computer vision: ResNet18~ \cite{heDeepResidualLearning2015} and TinyViT~\cite{wuTinyViTFastPretraining2022}.
 The \nmethods methods consist of 3 classical baseline methods Fine-tuning (FT), Gradient Ascent (GA), Successive Random Labels (SRL); 7 high-ranking methods in the NeurIPS'2023 Machine Unlearning competition~\cite{neurips-2023-machine-unlearning}, we name them Forget-Contrast-Strengthen (FCS), Masked-Small-Gradients (MSG), Confuse-Finetune-Weaken (CFW), Prune-Reinitialize-Match-Quantize (PRMQ), Convolution-Transpose (CT), Knowledge-Distillation-Entropy (KDE) and Repeated-Noise-Injection (RNI); and 8 recently published methods: Saliency Unlearning (SalUN)~\cite{fanSalUnEmpoweringMachine2024}, Catastrophic Forgetting-K (CF-k)~\cite{goelAdversarialEvaluationsInexact2023}, Exact Unlearning-K (EU-k)~\cite{goelAdversarialEvaluationsInexact2023}, SCalable Remembering and Unlearning unBound (SCRUB)~\cite{kurmanjiUnboundedMachineUnlearning2023}, Bad Teacher (BT)~\cite{chundawatCanBadTeaching2023}, Fisher Forgetting (FF)~\cite{golatkarEternalSunshineSpotless2020}, Influence Unlearning (IU)~\cite{kohUnderstandingBlackboxPredictions2017,izzoApproximateDataDeletion2021} and Negative Gradient Plus (NG+)~\cite{kurmanjiUnboundedMachineUnlearning2023}.
We evaluate the different unlearning methods based on four major aspects: privacy evaluation, performance retention, computational efficiency, and unlearning reliability.

The \textbf{contributions} of our study is to address the following research questions:

\begin{itemize}
    \item \textit{Q1.
Are the commonly used MU baselines reliable?} Most recently introduced MU methods are compared only with three baselines: Fine-tuning, Gradient Ascent, and Successive Random Labels, and not across recently introduced approaches.
We show that with proper hyperparameter selection, FT is a reliable baseline. In contrast, GA consistently performs poorly and should be replaced by the more recent Negative Gradient Plus (NG+)~\cite{kurmanjiUnboundedMachineUnlearning2023} that simultaneously reduces the performance in the forget set while maintaining the performance on the rest of the data points.

\item \textit{Q2.
How reliable are MU methods across datasets, models, and initializations?}  Our findings show that Masked-Small-Gradients is consistently among the best-performing methods across various metrics, unlike most recent MU methods.
In contrast, methods such as SRL do not consistently outperform others across different datasets.

\item \textit{Q3.
Which existing methods are the most reliable and accurate?}
Among \nmethods methods, our evaluation shows that certain methods, such as Masked-Small-Gradients (MSG), Convolution-Transpose (CT), and Fine-tuning (FT), exhibit desirable properties.
Specifically, MSG and CT show resilience against U-LiRA, which is a strong per-sample membership inference attack.
Furthermore, all three methods show consistency across datasets, and these methods could serve as reliable baselines for future studies.

\end{itemize}

{We publish the source code used for reproducing the experiments conducted in this paper at \url{https://github.com/xcadet/deepunlearn}}

\section{Related Works} \label{app:related}
\color{\mainchange}
We provide some context on data memorization in Machine Learning (\cref{section:related:memorization}), an overview of Machine Unlearning (MU) (\cref{app:mu}) with a focus on its application to Deep Neural Networks (DNNs) (\cref{app:mu_dnn}), the scenario of proceeding to unlearn after the model has been trained  (\cref{section:related:posthoc}), the link between Machine Unlearning and Differential Privacy (DP) (\cref{section:related:differential}), and existing benchmarks (\cref{section:related:benchmarks}).

\subsection{Data memorization in Machine Learning} \label{section:related:memorization}
Machine Learning methods aim to learn patterns from their training data. Nonetheless, multiple studies have raised awareness of Machine Learning models' capacity to leak information about their training datasets.

Ateniese et al. \cite{atenieseHackingSmartMachines2013} showed that one could use a trained machine learning classifier and obtain information about their training set. They reason that typical ML classifiers, such as Support Vector Machines (SVMs) and Hidden Markov Models (HMMs), must adapt their internal states to extract information from the training data.
They showed that trained classifiers could leak information about their training data.

Fredrikson et al. \cite{fredriksonPrivacyPharmacogeneticsEndEnd2014} raised awareness of the risk associated with Machine Learning models trained on sensitive data.
Their work studied the privacy risks associated with models trained to guide patients' medical treatment based on sensitive attributes. It showed that the trained model could infer sensitive attributes of patients.

Zhang et al. \cite{zhangUnderstandingDeepLearning2017a} showed that large neural networks could easily fit randomly labeled data.
Since random relabeling breaks the relationship between the image and its actual label, it implies that the effective capacity of large popular neural networks is sufficient for these models to memorize their training set. 

Carlini et al. \cite{carliniSecretSharerEvaluating2019} studied the case of generative sequence models that are sometimes trained on sensitive data such as private communication. They discussed the topic of unintended memorization, that is, neural networks can memorize information about the training data unrelated to the task they are trained for.
This finding raised significant privacy concerns, especially for models trained on sensitive information, as memorization could occur on data points considered outliers.
\color{black}

\subsection{Machine Unlearning} \label{app:mu}
Machine Unlearning is often first associated with the work from Cao et al.~\cite{caoMakingSystemsForget2015}, followed by Bourtoule et al.~\cite{bourtouleMachineUnlearning2021}, which proposes SISA (Sharded, Isolated, Sliced, Aggregated) as an exact unlearning method.
For a recent overview of MU, we refer to the survey from Xu et al.~\cite{xuMachineUnlearningSurvey2023}, which provides a taxonomy of common unlearning methods.
Furthermore, Zhang et al.~\cite{zhangReviewMachineUnlearning2023} review MU through privacy-preserving and security lenses.
The authors cover the Confidentiality, Integrity, and  Availability security triad and the need for Data Lineage, which relates to following the movement of data in a machine learning pipeline and understand from where it originates, where it is stored, and how it percolates in the system through transformation.
Some might have information on standard MU verification methods, privacy evaluation metrics, and datasets; we defer to the work of Nguyen et al.~\cite{nguyenSurveyMachineUnlearning2022} and Shaik et al.~\cite{shaikExploringLandscapeMachine2023}.

In the following, we focus on the MU taxonomy from Xu et al.~\cite{xuMachineUnlearningSurvey2023}, which considers Data Reorganization and Model Manipulation:

\subsubsection{Data Reorganization} methods focus on directly modifying the data to perform unlearning.
It is divided into Data Obfuscation, Data Pruning, and Data Replacement.
\begin{itemize}
    \item {\bf Data Obfuscation} refers to modifying the dataset to obscure the influence of the data to be unlearned: random relabeling and retraining~\cite{gravesAmnesiacMachineLearning2020}, SRL (Successive Random Labels), and Saliency Unlearning (SalUN).
    \item {\bf Data Pruning} usually relies on dividing the dataset into multiple sub-datasets and training sub-models on these subsets.
This is the category to which SISA~\cite{bourtouleMachineUnlearning2021} relates.
Our work does not consider methods associated with this setting as they make assumptions about the training process.

\item {\bf Data Replacement} attempts to unlearn by replacing the original dataset with transformed data that simplifies unlearning specific samples.
For instance, Cao et al.~\cite{caoMakingSystemsForget2015} replace the training data with summations of efficiently computable transformations.
Like data pruning, these methods tend to make strong assumptions about the training process.

\end{itemize}
\subsubsection{Model Manipulation} methods directly adjust the model parameters to remove the influence of specific data points.
Model manipulation is divided into Model Shifting, Model Replacement, and Model Pruning.

\begin{itemize}
    \item {\bf Model Shifting} directly updates the model parameters to offset the influence of the unlearned samples, such as using a single step of Newton's method on model parameters~\cite{guoCertifiedDataRemoval2020} or decremental updates \cite{schelterAmnesiaMachineLearning2020}, in our benchmark Fisher Forgetting (FF), Influence Unlearning (IU), and SalUN would represent these approaches.

    \item {\bf Model Replacement} uses pre-calculated parameters that do not reflect the data to forget to replace parts of the trained model.
For instance, when using decision trees, one can replace nodes affected by the forget set by pre-calculated nodes ~\cite{schelterHedgeCutMaintainingRandomised2021}.

These methods often make strong assumptions about the training process and the overall model.

\item {\bf Model Pruning} prunes specific parameters from the trained models to remove the influence of certain samples~\cite{jiaModelSparsityCan2024} or Prune-Reinitialize-Match-Quantize (PRMQ)~\cite{kgl_4_prmq} which prunes the model via L1 pruning, reinitializes parts of the model then train the model on $\mathcal{D}_R$.
\end{itemize}

\subsection{Machine Unlearning for Deep Neural Networks} \label{app:mu_dnn}
Initially, MU research primarily focused on linear models such as linear regression and logistic models.
Such models allow for the design of methods that assume the convexity of the loss function, rendering them less practical for DNN-based approaches.
Since DNNs can memorize parts of their training data, they are particularly relevant targets for MU, even more so when trained on large amounts of potentially personal data.
For Deep Learning models, unlearning raises additional challenges: 1) the non-convexity of the loss function of Deep Neural Networks \cite{choromanskaLossSurfacesMultilayer2015}, 2) the size of the models inducing large computational costs, 3) the randomness coming from the model's training process, such as the initialization seed, randomness in the mini-batch generation process, and 4) the fact that any model update impacts subsequent versions of the models, namely the weights at epoch $n + 1$ directly depend on the weights at update $n$.

When considering MU for DNN, Xu et al.~\cite{xuMachineUnlearningSurvey2023} notes that a standard scheme DNN is to focus only on the final layer, as it is expected for this layer to be the most relevant for the downstream task, and stems from the early works MU.
Nonetheless, Goel et al.~\cite{goelAdversarialEvaluationsInexact2023} showed that simply modifying the final layer is often insufficient to remove information related to $\mathcal{D}_f$.
However, other approaches, such as those from Golatkar et al.
~\cite{golatkarEternalSunshineSpotless2020,golatkarForgettingOutsideBox2020,golatkarMixedPrivacyForgettingDeep2021}, attempt to unlearn the full model via methods derived from Information Theory.
For instance, weight scrubbing on trained models can be done by approximating the Fisher information matrix.

\subsection{Post-Hoc Machine Unlearning} \label{section:related:posthoc}
While proactively designing deep-learning pipelines with built-in unlearning methods such as SISA can greatly simplify the unlearning process, many contemporary services relying on DNNs were not deployed with unlearning in mind.
This motivates searching for methods that can unlearn from already trained models without making assumptions about the training process.

Thus, we focus on {\em posthoc MU}, a scenario where we assume that the unlearning method is agnostic to the original training process of the model.
Under such a scenario, differences exist in terms of data availability at unlearning time.
For instance, whether one has access to the original training data $\mathcal{D}$,  the retain set $\mathcal{D}_R$, the forget set $\mathcal{D}_F$, or even some external set such as the validation set $\mathcal{D}_V$.
Therefore, careful consideration should be given to the data requirement associated with an unlearning method.
Indeed, some might require having access to both $\mathcal{D}_R$ and $\mathcal{D}_F$ at the unlearning time, while others assume that $\mathcal{D}_R$ is no longer available~\cite{chundawatZeroShotMachineUnlearning2023} making them more practical in real-world scenarios.
Throughout our benchmark, we make the same assumption as the NeurIPS2023 Unlearning Challenge \cite{neurips-2023-machine-unlearning}, where the unlearning methods had access to $f_O, \mathcal{D}_R, \mathcal{D}_F, \mathcal{D}_V$

\subsection{Machine Unlearning and Differential Privacy} \label{section:related:differential}
We based our Unlearning definition on Sekhari et al.~\cite{sekhariRememberWhatYou2021} and refer to their work on the distinction between Differential Privacy and the objective of Machine Unlearning.
In a high-level picture, differential privacy is a method for publicly sharing aggregated information about a population by describing the patterns discovered among the groups within the dataset while withholding specific information about individual data points.
A randomized algorithm $\mathcal{A}$ is $(\varepsilon, \delta)$-differentially private if for all datasets $D_1$ and $D_2$ that differ on a single data point, and all $S \subseteq \text{Range}(\mathcal{A})$,

\begin{equation}
\Pr[\mathcal{A}(D_1) \in S] \leq e^\varepsilon \cdot \Pr[\mathcal{A}(D_2) \in S] + \delta.
\end{equation}

In this definition, $\varepsilon$ (epsilon) is a non-negative parameter that measures the privacy loss, with smaller values indicating stronger privacy.
The parameter $\delta$ represents the probability of breaking differential privacy, ideally close to or equal to 0.

Despite enabling provable error guarantees for Unlearning methods, Differential Privacy requires strong model and algorithmic assumptions, making MU, derived from it, potentially less effective against practical adversaries \cite{jiaModelSparsityCan2024}.

\color{\mainchange}
\subsection{Machine Unlearning benchmarks} \label{section:related:benchmarks}

Grimes et al. \cite{grimesGoneNotForgotten2024} benchmark methods from the literature using CIFAR-100, a ResNet18 architecture, and an adapted version offline version U-LIRA metric. Their work investigates the effect of iterative unlearning, namely, unlearning occurs in multiple rounds to reflect multiple data unlearning queries.

Li et al. \cite{liWMDPBenchmarkMeasuring2024c} propose a benchmark for unlearning on Large Language Models. They suggest the Weapons of Mass Destruction Proxy (WMDP) dataset composed of $3,668$ multi-choice questions about biosecurity, cybersecurity, and chemical security. They aim to limit autoregressive Language models' ability to answer queries about hazardous knowledge while maintaining the model's ability to answer questions about nonhazardous knowledge. 

Cheng et al. \cite{chengMUBenchMultitaskMultimodal2024} propose a multi-task multimodal machine unlearning benchmark covering image, text, speech, and video modalities. They follow a retrain-free evaluation framework across nine datasets for different discriminative and generative tasks and modalities. In this work, we focus on image classification and consider the effect of model initializations and the methods at the top of the NeurIPS Machine Unlearning competition.

Ma et al. \cite{maDatasetBenchmarkCopyright2024} benchmarks Machine Unlearning with a focus on Copyright Infringement for Text-to-Image Diffusion models. They emphasize the ability of diffusion models to generate outputs that are closely similar to images from their training set.
In their work, they consider four viewpoints to address copyright infringement: style—the distinctive patterns of an artist—portrait—an individual's control over the use of their own portrait—artistic creation figures—for instance, animation characters are often protected by law—and licensed illustrations.

Zhang et al. \cite{zhangUnlearnCanvasStylizedImage2024a} benchmarks unlearning in Diffusion models. They proposed a dataset consisting of Seed images and stylized versions of these images and focused on evaluating Style and Object unlearning in Diffusion Models.

To our knowledge, Triantafillou et al. \cite{triantafillouAreWeMaking2024a}'s work is the most similar to ours. Their work leverages the results from the NeurIPS Unlearning Competition and evaluates both the methods from the competition and state-of-the-art methods.
We refer to them for additional discussion on the prize-winning methods of the competition.
Their work evaluates the methods on CASIA-SURF and FEMNIST using a ResNet-18 and searches for the hyper-parameters using a grid search on CASIA-SURF with three possible values per hyper-parameters
We consider five datasets, both Convolutional Neural Networks (CNN) and Vision Transformers (ViT), and perform the hyper-parameter search using the hyper-parameters optimizer framework Optuna \cite{akibaOptunaNextgenerationHyperparameter2019}.
Our findings indicate that the best-performing method from the competition is less reliable when tested across different architectures and datasets.
Given that unlearning methods can be sensitive to hyper-parameter choices, our benchmark ensures fair comparisons by searching for 300 hyper-parameter configurations (three searches of 100 combinations each), with each search starting from a model with different initial weights.
\color{black}

\section{Background of Machine Unlearning}\label{unlearning_taxonomy}

\subsection{Problem setting}
Starting with a \textbf{training set} $\mathcal{D} = \{(\mathbf{x}_i, y_i)\}_{i=1}^{N}$ and a trained model $f_O$, referred to as the \textbf{original model}, the objective of a MU method $U$ is to \textit{remove} the influence of a particular subset of training set $\mathcal{D}_F = \{(\mathbf{x}_i, y_i)\}_{i=1}^{K} \subset \mathcal{D}$ referred to as the \textbf{forget set} where $K \ll N$. The rest of the training set $\mathcal{D}_R = \mathcal{D}\backslash \mathcal{D}_F$ is called the \textbf{retain set}. The forget and retain sets are distinct and complementary subsets of the training set. The outcome of MU is an  \textbf{unlearned} model $f_U$, the aim for which is to {\em perform} on par with a model {\em retrained from scratch} on $\mathcal{D}_R$; this latter model $f_R$ is referred to as the \textbf{retrained} model. 

We denote the weights of the original model and retrained models' weights as $\theta_O$ and $\theta_R$, respectively. For evaluations, we consider two held-out sets: the \textbf{validation set} $\mathcal{D}_V$ and \textbf{test set} $\mathcal{D}_T$, both drawn from the same distribution as $\mathcal{D}$. We consider the accuracy of the retrained model to be the optimal accuracy.
One critical assumption we make is that the MU method has access to $\theta_O$, $\mathcal{D}_R$, $\mathcal{D}_F$, and $\mathcal{D}_V$.

\subsection{Unlearning Evaluation.} To evaluate the success of unlearning, one approach is to check whether data points in $\mathcal{D}_F$ still influence the predictions made by the unlearned model ~\cite{carliniMembershipInferenceAttacks2022,kurmanjiUnboundedMachineUnlearning2023,hayesInexactUnlearningNeeds2024}. This is commonly done via {\em influence functions}~\cite{kohUnderstandingBlackboxPredictions2017,basuInfluenceFunctionsDeep2021,grosseStudyingLargeLanguage2023} and membership inference attacks~(MIA)~\cite{shokriMembershipInferenceAttacks2017}.
MIA has become one of the most common approaches for evaluating MU methods.
It aims to determine whether specific data points were part of the original training dataset based on the unlearned model. 

\subsubsection{Indiscernibility}
We compute the population-based U-MIA, denoted with $\text{MIA}(\mathcal{D}, f_U)$ evaluated on data $\mathcal{D}$\footnote{The methodology for this follows the classic MIA: we compute the losses on $\mathcal{D}_F$ and $\mathcal{D}_V$, we shuffle and trim them so that they are of equal size. We then train logistic regression models in a 10-fold cross-validation and compute the average accuracy across the folds.}. Using this we define the \textbf{discernibility}\label{metric:discernibility} metric as $\text{Disc}(\mathcal{D}_V, f_U) = |2 \times \text{MIA}(\mathcal{D}, f_U) - 1| \in [0, 1]$, and similarly, \textbf{indiscernibility} \label{metric:indiscernibility} is given by $\text{Indisc}(\mathcal{D}, f_U) = 1 - \text{Disc}(\mathcal{D}, f_U) \in [0, 1]$. We set $\mathcal{D}$ to either the test set $\mathcal{D}_T$ or validation set $\mathcal{D}_V$. 
The indiscernibility equals $1$ when the accuracy of the MIA is not better than random guessing.

A more recent MIA variation is the unlearning likelihood ratio attack (U-LiRA)~\cite{kurmanjiUnboundedMachineUnlearning2023}. 
As highlighted by \cite{hayesInexactUnlearningNeeds2024}, U-LiRA is a more robust evaluation approach for approximate MU. Nonetheless, U-LiRA is much more computationally demanding than U-MIA. We evaluate the methods that defeat the weaker, less expensive U-MIA attack, additionally against the U-LiRA attack.

\subsubsection{Accuracy} The classification accuracy of the unlearned model on the retain set should be as close as possible to that of the original model. First, we consider three metrics derived directly from the model's classification accuracy on different sets: \textbf{retain accuracy}~(RA), \textbf{forget accuracy}~(FA), and \textbf{test accuracy}~(TA). RA is defined as 
 \begin{equation}
    \text{RA}(\mathcal{D}_R, f_U) = \frac{1}{|\mathcal{D}_R|}\sum_{(\mathbf{x}_i, y_i) \in \mathcal{D}_R}\mathbf{1}_{y_i = f_U(\mathbf{x}_i)} \in [0, 1].
 \end{equation}
The metrics for FA and TA can be derived by replacing $\mathcal{D}_R$ with $\mathcal{D}_F$ and $\mathcal{D}_T$, respectively. Second, while RA, FA, and TA give us insight into the overall accuracy of the unlearned model, they do not capture how well it performs compared to a retrained model $f_R$.
Considering the $f_R$ model as a gold standard, we derive three more metrics: \textbf{retain retention}~(RR), \textbf{forget retention}~(FR), \textbf{test retention}~(TR), where RR is given by,
\begin{equation}
    \text{RR}(f_U, f_R) = \frac{\text{RA}(\mathcal{D}_R, f_U)}{\text{RA}(\mathcal{D}_R, f_R)} \in [0, +\infty),
\end{equation}
and formulas for FR and TR can be derived using FA and TA, respectively. An unlearned model with a score of $1$ indicates that its accuracy perfectly matches the accuracy of the reference retrained model.
A score below $1$ indicates that the model underperforms, and a score above $1$ indicates that the model overperforms.

From these, we can define deviation scores, that provide information on the divergence of the unlearned model from the retrained model's performance.
Namely we consider the \textit{retain set deviation} $DR$ (\cref{equation:retain_deviation}),  \textit{forget set deviation} $DF$ (\cref{equation:forget_deviation}), and \textit{test set deviation} $DT$ (\cref{equation:test_deviation}).
\begin{equation} \label{equation:retain_deviation}
  DR = |RR(f_U, f_R) - 1| \in [0, +\infty)
\end{equation}
\begin{equation} \label{equation:forget_deviation}
  DF = |FR(f_U, f_R) - 1| \in [0, +\infty)
\end{equation}
\begin{equation} \label{equation:test_deviation}
  DT = |TR(f_U, f_R) - 1| \in [0, +\infty)
\end{equation}
We further define the \textbf{Retention Deviation}~(RetDev) as:
\begin{equation}
    \text{RetDev} = DR + DF + DT \in [0, +\infty),
\end{equation}
Which provides information on the cumulative divergence of the unlearned model regarding retention score. The closer to 0, the better, as 0 indicates that the model perfectly matches the performance of the retrained model.

\subsubsection{Efficiency} \label{metric:rte} run-time efficiency~(RTE) of a MU method should ideally be higher than the naive approach of just retraining the model from scratch on the retain set. As the high computational cost of the retraining algorithm motivated the development of approximate MU methods, we evaluate how much faster each MU method is compared to the retraining algorithm. 

We define RT of $U$ as the number of seconds it takes to complete, denoted as $\text{RT}(U)$ (we use the same machine and resources for all the experiments). To indicate the relative speedup compared to the retrained model, we define the RTE of an unlearn method $U$ as:
\begin{equation}
\frac{\text{RT}(U_R)}{\text{RT}(U)} \in [0, +\infty),
\end{equation}
Where $U_R$ denotes the retrain method. The RTE of retraining from scratch is 1; any method with an RTE of less than 1 is slower than retraining from scratch, and vice versa.

\section{Machine Unlearning Methods} \label{sec:unlearn_methods}
Here, we discuss the main unlearning methods considered. 
First, we cover the classical baselines (\cref{section:unlearn_methods:classical}).
Then, methods from the competition and literature (\cref{section:unlearn_methods:sota})
\subsection{Classical Baselines} \label{section:unlearn_methods:classical}
\textbf{FineTune (FT)} keeps training the original model $f_O$ only on the retain set $\mathcal{D}_R$ for several epochs; the number of epochs is considered as a hyper-parameter and varies from one dataset and architecture. The models are trained by minimizing the cross-entropy over the retain set.

\textbf{Successive Random Labels (SRL)} trains $f_O$ on both the forget set $\mathcal{D}_F$ and $\mathcal{D}_R$ while the labels of $\mathcal{D}_F$ are randomly assigned redrawn from a uniform distribution $\mathcal{U}_C$, with $C$ the number of classes at each epoch.

\textbf{Gradient Ascent (GA)} trains the model using gradient \emph{ascent} steps on the $\mathcal{D}_F$. Namely, we compute the cross-entropy loss for a batch and take a step toward the gradient rather than opposite it.

\subsection{State-of-the-art MU methods} \label{section:unlearn_methods:sota}
In addition to the classical baselines, we evaluate 15 recent MU methods.
First, we discuss the seven top-performing methods from the competition.
Then, we cover eight methods from the literature.
We first discuss the seven top-performing methods from the Machine Unlearning Competition 2023 on Kaggle. 

\begin{table}[h]
    \centering
     \caption{Kaggle - Leaderboard and their associated names in this work.}
    \label{table:kaggle_ranking_and_names}
    \begin{tabular}{lcc}
        \toprule
         Kaggle Rank & Name & Acronym \\
         \midrule
         1 & Forget-Contrast-Strengthen & (FCS) \\ 
         2 & Masked-Small-Gradients & (MSG)\\ 
         3 & Confuse-Finetune-Weaken & (CFW)\\ 
         4 & Prune-Reinitialize-Match-Quantize & (PRMQ)\\ 
         5 & Convolution-Transpose & (CT)\\ 
         6 & Knowledge-Distillation-Entropy & (KDE) \\ 
         7 & Repeated-Noise-Injection& (RNI)\\ 
         \bottomrule
    \end{tabular}
\end{table}

\textbf{Forget-Contrast-Strengthen (FCS)} \cite{kgl_1_fcs} minimizes the \acrfull{kld} between the model's output on $\mathcal{D}_F$ and uniform distribution over the output classes, then alternatively optimizes a contrastive loss between the model's outputs on $\mathcal{D}_R$ and $\mathcal{D}_F$, and minimizes the cross-entropy loss on $\mathcal{D}_R$. 

\textbf{Masked-Small-Gradients (MSG)}\cite{kgl_2_msg} accumulates gradients via gradient \textit{descent} on the $\mathcal{D}_R$ and gradient \textit{ascent} on the $\mathcal{D}_F$, then reinitialize weights with the smallest absolute gradients while dampening subsequent weights updates on the $\mathcal{D}_R$ for the other weights. 

\textbf{Confuse-Finetune-Weaken (CFW)}\cite{kgl_3_cfw} injects noise into the convolutional layers and then trains the model using a class-weighted cross-entropy on $\mathcal{D}_R$, then injects noise again toward the final epochs.

 \textbf{Prune-Reinitialize-Match-Quantize (PRMQ)}~\cite{kgl_4_prmq} first prunes the model via L1 pruning, reinitializes parts of the model, optimizes it using a combination of cross-entropy and a mean-squared-error on the entropy between the outputs of $f_O$ and $f_U$ on $\mathcal{D}_R$ and finally converts $f_U$'s weights to half-precision floats.
 
\textbf{Convolution-Transpose} \cite{kgl_5_ct} simply transposes the weights in the convolutional layers and trains on $\mathcal{D}_R$.

\textbf{Knowledge-Distillation-Entropy (KDE)}~\cite{kgl_6_kde} uses a teacher-student setup. Both student and teacher start as copies of the original model, and then the student's first and last layers are reinitialized. The student $f_U$ minimizes its \acrfull{kld} with the $f_O$ over $\mathcal{D}_V$, then  minimizes
a combination of losses: a soft cross-entropy loss between $f_U$ and $f_O$, a cross-entropy loss on outputs of $\mathcal{D}_R$ from $f_U$, and the \acrshort{kld} between $f_U$ and $f_O$ on $\mathcal{D}_R$.

\textbf{Repeated-Noise-Injection (RNI)}~\cite{kgl_7_rni} first reinitializes the final layer of the model, then repeatedly injects noise in different layers of the model while training on the $\mathcal{D}_R$.

We further consider eight state-of-the-art methods introduced in the literature. 

\textbf{Fisher Forgetting (FF)}~\cite{golatkarEternalSunshineSpotless2020,fanSalUnEmpoweringMachine2024} adds noise to $f_O$ with zero mean and covariance determined by the 4th root of Fisher Information matrix with respect to  $\theta_O$ on $\mathcal{D}_R$.

\textbf{Influence Unlearning (IU) }~\cite{izzoApproximateDataDeletion2021,warneckeMachineUnlearningFeatures2023,jiaModelSparsityCan2024} uses Influence Functions\cite{cookResidualsInfluenceRegression1982} to determine the change in $\theta_O$ if a training point is removed from the training loss. IU estimates the change in model parameters from $\theta_O$ to the model trained without a given data point. We use the first-order WoodFisher-based approximation from \cite{jiaModelSparsityCan2024}.

\textbf{Catastrophic Forgetting - K (CF-K)}~\cite{goelAdversarialEvaluationsInexact2023} freezes the first layers then trains the last $k$ layers of the model on $\mathcal{D}_R$.

\textbf{Exact Unlearning - K (EU-K)}~\cite{goelAdversarialEvaluationsInexact2023} freezes the first layers and then restores the weights of the last $k$ layers to their initialization state. We randomly reinitialize the weights instead so that the method no longer requires knowledge about the training process of $f_O$.

\textbf{\acrfull{unl_scrub}}~\cite{kurmanjiUnboundedMachineUnlearning2023} leverages a student-teacher setup where the model is optimized for three objectives: matching the teacher's output distribution on $\mathcal{D}_R$, correctly predicting the $\mathcal{D}_R$ set and ensuring the output distributions of the teacher and student diverge on the $\mathcal{D}_F$

\textbf{\acrfull{unl_salun}}~\cite{fanSalUnEmpoweringMachine2024} 
determines via gradient \textit{ascent} which weights of $\theta_O$ are the most relevant to $\mathcal{D}_F$, then trains the model simultaneously on $\mathcal{D}_R$ and $\mathcal{D}_F$  with random labels on $\mathcal{D}_F$, while dampening the gradient propagation based on the selected weights.

\textbf{Negative Gradient Plus (NG+) }~\cite{kurmanjiUnboundedMachineUnlearning2023} is an extension of the Gradient Ascent approach where additionally a gradient descent step is taken over the $\mathcal{D}_R$.

\textbf{Bad Teacher (BT)}~\cite{chundawatCanBadTeaching2023} uses a teacher-student approach with two teachers: the original model and a randomly initialized model - the bad teacher-, the student starts as a copy of $f_U$ then learns to mimic the $f_O$ on $\mathcal{D}_R$ and the bad teacher on the $\mathcal{D}_F$.

\section{Experimental Evaluation}\label{sec_exp_eval}

{\bf Experiments.} We evaluate the \nmethods recent MU methods (described in \cref{sec:unlearn_methods}) across \ndatasets benchmark datasets: MNIST~\cite{lecunGradientbasedLearningApplied1998}, FashionMNIST~\cite{xiaoFashionMNISTNovelImage2017}, CIFAR-10~\cite{krizhevskyLearningMultipleLayers2009}, CIFAR-100~\cite{krizhevskyLearningMultipleLayers2009}, and UTK-Face~\cite{zhangAgeProgressionRegression2017}.
These datasets vary in difficulty, number of classes, instances per class, and image sizes. We consider two model architectures: a TinyViT and a ResNet18 model.
These datasets are common to the Computer Vision community and readily available to the community in Pytorch, apart from UTK-Face.
Hence, in total, we evaluate nine different combinations of models and architectures: ResNet18 and TinyViT on MNIST, FashionMNIST, CIFAR-10, CIFAR-100, and ResNet18 on UTKFace.
More information on the data sets, hyperparameters, and data augmentations used to train the original and retrained models is provided in \cref{app:datasets}.

\color{\mainchange}
\subsection{Data splits}
We use a fixed seed $s=123$ to generate the different data splits.
If a Training and Test split exists for every dataset, we keep it and refer to the original training split as the development split.
Otherwise, we divide the dataset following a $80\%/20\%$ (Development / Test) split. The Development set is then further divided. We use the Development split for training the model, defining the forget and retain set, while we use the Test set to evaluate the final performance of each method.
We split the Development into Training $\mathcal{D}$ and Validation $\mathcal{D}_V$ splits following a $80\%/20\%$ split.
The Training set is used to train the Original model, while the validation set is used for the hyper-parameters searches.
We subdivide the Training set into the Retain $\mathcal{D}_R$ and Forget $\mathcal{D}_F$ sets following a $90\%/10\%$ split.
As such, we consider a $10\%$ forgetting budget, which is common in the literature, and draw these $10\%$ from the Training set following a Uniform distribution.
\color{black}

\subsection{Datasets} \label{app:datasets}

\begin{table*}\label{table:dataset_information}
\caption{Summary of the datasets used in the benchmark, with split sizes, number of classes per dataset, task and color information.}
\resizebox{\textwidth}{!}{%
\begin{tabular}{lcccccr}
\toprule
Dataset & Retain size & Forget size & Validation size & Test size & Number of classes & Task \\
\midrule
MNIST & 45,900 & 5,100 & 9,000 & 10,000 & 10 & Handwritten digit classification (Grayscale) \\
FashionMNIST & 45,900 & 5,100 & 9,000 & 10,000 & 10 & Fashion item classification (Grayscale) \\
CIFAR-10 & 38,250 & 4,250 & 7,500 & 10,000 & 10 & Object classification (RGB) \\
CIFAR-100 & 38,250 & 4,250 & 7,500 & 10,000 & 100 & Fine-grained object classification (RGB) \\
UTKFace & 14,508 & 1,613 & 2,846 & 4,741 & 5 & Age classification (RGB) \\
\bottomrule
\end{tabular}
}
\end{table*}

\subsubsection{CIFAR-10}
is a widely used dataset in computer vision and machine learning. It comprises 60,000 32x32 color images in 10 different classes, with 6,000 images per class.
The dataset is divided into 50,000 training images and 10,000 testing images.
CIFAR-10 represents a diverse range of everyday objects, such as airplanes, automobiles, birds, and cats, making it a challenging task for image classification.
The simplicity of the images combined with the variety of categories makes CIFAR-10 a suitable dataset to test the efficacy of machine unlearning algorithms in effectively unlearning information without compromising the model's performance on the remaining data.

{\em Data Augmentations:} random cropping to 32x32 with 4-pixel padding, 50\% random horizontal flipping, and per-channel normalization with a mean of $[0.4919, 0.4822, 0.4465]$ and standard deviation of $[0.2023, 0.1994, 0.2010]$.
At test time, we resize to 32x32 and normalize.

\subsubsection{CIFAR-100}
is a more complex extension of CIFAR-10, containing 100 classes with 600 images per class, split into 500 training images and 100 testing images per class. Each class is labeled with a "fine" label and grouped into 20 "coarse" labels, adding another layer of classification difficulty. The increased number of classes and finer granularity make CIFAR-100 an intriguing dataset for machine unlearning benchmarks. It poses a more significant challenge for models to forget specific classes or groups while retaining knowledge of others, thus testing the unlearning algorithms' precision and effectiveness in handling more granular and complex datasets.

{\em Data Augmentations:} random cropping to 32x32 with 4-pixel padding, 50\% random horizontal flipping, and per-channel normalization with a mean of $[0.5071, 0.4865, 0.4409]$ and standard deviation of $[0.2673, 0.2564, 0.2762]$.
At test time, we resize to 32x32 and normalize.

\subsubsection{MNIST}
is a well-known benchmark in handwritten digit recognition. It comprises 70,000 grayscale images of handwritten digits (0-9), 60,000 used for training, and 10,000 for testing.
Each image is 28x28 pixels in size.
We consider MNIST due to its simplicity and extensive research and development history.
The simplicity of MNIST allows researchers to focus on the fundamental aspects of unlearning techniques without the additional complexity introduced by color or high resolution, providing a clear assessment of the effectiveness of unlearning algorithms in a controlled setting.

{\em Data Augmentations:} conversion to 3 channels, resizing to 32x32 such that both ResNet18 and TinyViT use the same input resolution, 50\% random horizontal flipping, and per-channel normalization with a mean of $[0.1307, 0.1307, 0.1307]$ and standard deviation of $[0.3081, 0.3081, 0.3081]$.
We convert to 3 channels at test time, resize to 32x32, and normalize.

\subsubsection{Fashion MNIST}
is a more challenging replacement for MNIST. It contains 70,000 grayscale images of fashion items in 10 categories: shirts, trousers, and sneakers.
Like MNIST, each image is 28x28 pixels, but the increased complexity and variability of clothing items make it a more challenging classification task.
Fashion MNIST provides a more realistic and intricate dataset than MNIST, testing the unlearning algorithms' ability to handle real-world-like variability and ensuring that they can effectively remove learned information while maintaining performance on a moderately complex dataset.

{\em Data Augmentations:} conversion to 3 channels, resizing to 32x32, 50\% random horizontal flipping, and per-channel normalization with a mean of $[0.2860, 0.2860, 0.2860]$ and standard deviation of $[0.3560, 0.3560, 0.3560]$.
We convert to 3 channels at test time, resize to 32x32, and normalize.

\subsubsection{UTKFace}
UTKFace is a large-scale face dataset containing over 20,000 images of faces with annotations of age, gender, and ethnicity.
The images vary in size and cover a wide range of ages, from 0 to 116. UTKFace is particularly interesting due to the sensitive nature of the data and the need for privacy-preserving techniques.

{\em Data Augmentations:} resizing to 224x224, and per-channel normalization with a mean of $[0.485, 0.456, 0.406]$ and standard deviation of $[0.229, 0.224, 0.225]$.
We apply the same transformation at test time.

For each dataset, the Original and Retrained models are trained using the same hyper-parameters (provided in Table \ref{tab:summary})

\subsection{Neural Network Architectures} \label{app:models}
We consider two families, ResNet (Residual Network) \cite{heDeepResidualLearning2015} and ViT (Vision Transformer) \cite{dosovitskiyImageWorth16x162021}, which are prominent architectures in computer vision.
We consider ResNet18 and a TinyViT\cite{wuTinyViTFastPretraining2022} with approximately 11M learnable parameters for a fair comparison between two fundamentally different architectures.
This provides insights into how architectural differences impact the unlearning process and helps understand the trade-offs between convolutional and transformer-based models regarding reliability and computational efficiency.

\subsubsection{ResNet: ResNet18}
 Introduced by He et al.~\cite{heDeepResidualLearning2015}, it facilitates the training of deep networks through shortcut connections, which mitigates the problem of vanishing gradients. The ResNet18 is known for its balance between performance and computational efficiency.

 \subsubsection{Vision Transformer (ViT): TinyViT}
 Introduced by Dosovitskiy et al.~\cite{dosovitskiyImageWorth16x162021}, adapts the transformer architecture to image classification by treating images as sequences of patches.
 We consider TinyViT from Wu et al.~\cite{wuTinyViTFastPretraining2022}, as it is a compact version of ViT designed to be parameter-efficient while maintaining high performance.

The performance of the MU methods can change across datasets, model configurations, and model initializations; a reliable MU method remains consistent across these changes. For each method, dataset and model combination, we unlearn from Original models initialized using 10 different seeds and consider the average performance across seeds. 

\subsection{Hyper-parameters search} \label{section:methodology:hyper_parameters}

A further observation is that prior research tends to compare MU methods with default hyperparameters, potentially leading to a less competitive performance of the method.
To ensure that each method performs optimally, we perform three hyperparameter sweeps to find the best set of hyperparameters for each method. We use the same number of searches for each method to ensure a fair comparison. 
Each hyper-parameter sweep uses 100 trials to minimize four loss functions: Retain Loss ($\mathcal{L}_{\text{Retain}}$ equation \ref{equation:loss_retain}), Forget Loss ($\mathcal{L}_{\text{Forget}}$ equation \ref{equation:loss_forget}), Val Loss ($\mathcal{L}_{Val}$ equation \ref{equation:loss_validation}), and Val MIA ($\mathcal{L}_{\text{Val-MIA}}$ equation \ref{equation:loss_discernibility}) given by
\begin{equation} \label{equation:loss_retain}
    \mathcal{L}_{Retain} = \alpha\times|\text{RA}(f_U) - \text{RA}(f_R)|
\end{equation}
\begin{equation} \label{equation:loss_forget}
    \mathcal{L}_{Forget} = \beta\times|\text{FA}(f_U) - \text{FA}(f_R)|
\end{equation}
\begin{equation} \label{equation:loss_validation}
    \mathcal{L}_{Val}  = \gamma\times|\text{VA}(f_U) - \text{VA}(f_R)|
\end{equation}
\begin{equation} \label{equation:loss_discernibility}
    \mathcal{L}_{\text{Val-MIA}} = \eta \times \text{Disc}(\mathcal{D}_V, f_U)
\end{equation}

Where the $\mathcal{L}_{\text{Retain}}$  captures the divergence in accuracy between the retrained and unlearned model over the $\mathcal{D}_R$, $\mathcal{L}_{\text{Forget}}$ and $\mathcal{L}_{Val}$ capture the divergence over $\mathcal{D}_F, \mathcal{D}_V$ respectively and $\mathcal{L}_{\text{Val-MIA}}$ captures whether the loss distributions over $\mathcal{D}_F$ and $\mathcal{D}_V$ are distinguishable from one another via the discernibility score defined in Section \ref{metric:discernibility}.
We set $\alpha=\beta=\gamma=\frac{1}{3}$ and $\eta=1$ as we found these values to balance the importance of importance retention and resilience to Membership Inference Attacks. Per unlearning method, we use the hyperparameter configuration that minimizes the four loss terms when evaluating the method.
Thus, for each unlearning method, we first unlearn $300$ models to do the hyper-parameter sweep and then unlearn $10$ models with the best set of hyper-parameters, which amounts to $5,580$ per dataset for a given architecture and $50,220$ for the 9 dataset/model combinations.

\subsection{Privacy Evaluation} \label{app:mias}
\subsubsection{Unlearning-Membership Inference Attack (U-MIA)}
A common approach to evaluate the quality of unlearning methods is to attack the unlearned models with a form of Membership inference Attack (MIA).
Membership Inference Attacks attempt to determine whether a specific data point was part of the model train data. 
The efficacy of the Membership Inference Attack has been used as a metric to evaluate the success of unlearning algorithms.
A general approach to such an attack is as follows. 
Assume $f\theta$ is a trained model with parameters $\theta$, and let $\mathcal{L}$ be a loss function, such as the cross-entropy loss.
Then, compute the losses for each sample from two sets of data $A$ and $B$ (of equal size) and train a binary classification model such as logistic regression with labels $y^A_i = 1$ for points $i$ in $A$ and $y^B_i = 0$ for points $i$ in $B$.
An accuracy score from the classifier close to $1.0$ indicates that the classifier can perfectly distinguish between samples from $A$ and $B$ based on the loss values.
A score of $0.5$ indicates that the ability to distinguish is close to random. 

\subsubsection{Unlearning-Likelihood Ratio Attack (U-LiRA)}
The performance of a general MIA can be improved by considering, \textit{e.g.}, a per-sample attack such as LiRA \cite{carliniMembershipInferenceAttacks2022,hayesInexactUnlearningNeeds2024}.
For any given point, we wish to determine whether the outputs from the unlearned models differ from those of models that have never seen the data point.
To assess the attack robustly, we evaluate it across multiple models, using shadow models trained on various retain/forget sets. Specifically, we first train $n$ models based on $n$ splits of the training data. This train data is then split into 10 random retain and forget splits; hence, we unlearn a total of $10n$ models.
We then perform hyper-parameter sweeps, similar to what we do in the original results and unlearn using the optimal hyper-parameters, except that we consider $\frac{n}{2}$ sweeps and conduct $200$ trials per sweep to determine the best hyper-parameters.
In our setting, we set $n=64$.

\begin{table*}[t!] 
\centering
\caption{Ranking by performance on Retention Deviation and Indiscernibility across datasets and architectures. We count the number of times each method appears in the Best Performers group (G1), Average performance group (G2) and Worst performers group (G3) (see \S\ref{sec_exp_eval}). The final rank is computed based on the number of times the method appears in G1 with occurrences in G2 and G3 used to break ties if needed. If a method does not produce any usable models, it is assigned to a Failed group (F). Three methods appear in the top 3 for both performance measures: MSG (1st and 1st), CT (3rd and 1st) and KDE (3rd and 2nd).}
\label{table:overall_ranking}
\centering
\begin{tabular}{llcccccllcccccllc}
\toprule
& & \multicolumn{4}{c}{Retention Deviation} & & & & \multicolumn{4}{c}{Indiscernibility} & & & & Pareto Frontier \\
\cmidrule(r){3-6} \cmidrule(r){10-13} \cmidrule{17-17}
Rank & Method & G1 & G2 & G3 & F & & Rank & Method & G1 & G2 & G3 & F & & Rank & Method  & Times on Frontier \\
\midrule
1 & FT & 8 & 1 & 0 & 0 & & 1 & CT & 9 & 0 & 0 & 0     & & 1 & CFW  & 4 \\
1 & MSG & 8 & 1 & 0 & 0 & & 1 & MSG & 9 & 0 & 0 & 0   & & 2 & CT   & 3 \\
2 & PRMQ & 7 & 2 & 0 & 0 & & 2 & CFW & 7 & 2 & 0 & 0  & & 2 & FT   & 3 \\
3 & CT & 7 & 1 & 1 & 0 & & 2 & RNI & 7 & 2 & 0 & 0    & & 3 & PRMQ & 2 \\
\cmidrule(r){15-17}
3 & KDE & 7 & 1 & 1 & 0 & & 2 & KDE & 7 & 2 & 0 & 0   & & 4 & KDE & 1 \\
3 & CFW & 7 & 1 & 1 & 0 & & 3 & FT & 6 & 3 & 0 & 0    & & 4 & FCS & 1 \\
\cmidrule(r){1-6}
4 & FCS & 6 & 3 & 0 & 0 & & 3 & PRMQ & 6 & 3 & 0 & 0  & & 4 & SCRUB & 1 \\
4 & SalUN & 6 & 3 & 0 & 0 & & 3 & SalUN & 6 & 3 & 0 & 0 & & 4 & SRL & 1\\
\cmidrule(r){8-13}
5 & NG+ & 5 & 4 & 0 & 0 & & 4 & SRL & 6 & 2 & 1 & 0 & & 4 & CF-k & 1\\
5 & SRL & 5 & 4 & 0 & 0 & & 5 & NG+ & 5 & 4 & 0 & 0  & & 5 & BT & 0\\
6 & SCRUB & 4 & 3 & 1 & 1 & & 5 & FCS & 5 & 4 & 0 & 0 & & 5 & GA & 0\\
7 & BT & 2 & 7 & 0 & 0 & & 6 & SCRUB & 5 & 3 & 0 & 1  & & 5 & MSG & 0\\
7 & RNI & 2 & 7 & 0 & 0 & & 7 & BT & 4 & 5 & 0 & 0 &  & 5 & NG+ & 0\\
8 & CF-k & 2 & 3 & 2 & 2 & & 8 & CF-k & 1 & 2 & 4 & 2 & & 5 & RNI & 0 \\
9 & IU & 1 & 0 & 2 & 6 & & 9 & EU-k & 1 & 2 & 2 & 4 & & 5 & SalUN & 0\\
10 & EU-k & 0 & 5 & 0 & 4 & & 10 & GA & 0 & 4 & 4 & 1 & & 5 & EU-k & 0 \\
11 & GA & 0 & 1 & 7 & 1 & & 11 & IU & 0 & 0 & 3 & 6 & & 5 & IU & 0 \\
12 & FF & 0 & 0 & 0 & 9 & & 12 & FF & 0 & 0 & 0 & 9 & & 5 & FF & 0 \\
\bottomrule
\end{tabular}
\end{table*}

\subsection{Ranking the methods}
A challenge in comparing MU method performance comes from the potential proximity of the evaluation metrics.
As a simple example, suppose we have four methods $U_1,...,U_4$ with accuracies: 98\%, 99\%, 50\%, 1\%, respectively; if we simply rank the methods, the rank itself would not be representative of the fact that, e.g., $U_1$ and $U_2$ are much above $U_3$ and $U_4$. To enable distinctions based on proximities, we use Agglomerative Clustering and define cut-off points such that we obtain three clusters: (1) Best performers (G1), (2) Average performers (G2), and (3) Worst performers (G3). 
If a method does not produce ten usable models, one per original model, we assign it to the Failed group (F).
For each method, we count the times it appears in the three groups (with nine being the maximum). To obtain a final ranking of the methods, we first rank them using the number of times they appear in the Best Performers group (G1); if ties occur, we break them with the Average Performers (G2) group. If ties persist, the Worst Performers (G3) group is the final tie-breaker. This method ensures a clear and fair ranking by considering each performance group in order of importance.

\color{\mainchange}
\subsubsection{Pareto Frontier}
We rank methods based on the number of times they lie on the Pareto Frontier.
The Pareto frontier represents the set of methods that are non-dominated, that is, methods such that no other method performs better on both metrics simultaneously.

To determine this, we consider two metrics: (1) The Performance Retention Deviation metric to be minimized and (2) The Indiscernibility to be maximized.

To identify the Pareto frontier, we proceed as follows:
\begin{enumerate}
    \item  We gather the values of the two metrics for each method.
    \item We sort the methods by the first metric in ascending order (since it needs to be minimized) and secondarily by the second metric in descending order (since it needs to be maximized).
    \item We iterate through the sorted methods and retain only those not dominated by any other.
    \item The remaining methods after this filtering step constituted the Pareto frontier, representing the optimal trade-off between the two metrics, where improving one metric would necessarily worsen the other.
\end{enumerate}

\subsection{Consistent Unforgettability}
We question two aspects: 
\begin{enumerate}
    \item Are there images for which the U-MIA is consistently successful across model seeds?
    \item Are there images that multiple methods fail to unlearn?
\end{enumerate}
To answer these questions, we perform a 10-fold cross-validation U-MIA.
Namely, given a model $f_\theta$ and two sets of equal size, one containing only the losses associated with images from the Forget set $\mathcal{D}_f$, and the other containing only losses associated with images from the Test set $\mathcal{D}_T$, we take the union of these sets and generate $10$ subset in direct sum. We then train $10$ Logistic Regression classifiers, each using $9$ subsets to predict the remaining.
We keep track of each prediction made by the $10$ models, allowing us to get predictions for the entire Forget set.
We apply each unlearning method to the Original model resulting from each of the $10$ initialization seeds. 
Since the forget set is the same across all experiments, we can now determine how many times each of the elements of $\mathcal{D}_f$'s membership is correctly inferred.
From these, we can answer both questions, first by analyzing across seeds and then by analyzing across unlearning methods.
\color{black}

\section{Main Results}
We ranked the MU methods based on three criteria: Performance Retention Deviation, Indiscernibility, and the number of times they lie on the Pareto Frontier (\cref{table:overall_ranking}).
We provide the performance table for CIFAR-100 using ResNet-18 (\cref{tab:comp_res_cifar100}).
Since the groups are obtained using a clustering algorithm, they do not convey information about the distance between groups; as such, we provide such information for FashionMNIST and CIFAR-100 (\cref{table:group_distance}).
We gathered the Run-Time Efficiency of each method across datasets for the ResNet-18 architecture (\cref{table:rte_resnet18})
We provide we provide per dataset and architecture results in the Appendix.

\begin{table*}
\begin{center}
    
\caption{CIFAR-100 - ResNet18. CIFAR-100 provides the most visible comparison as there is a large gap in performance between the Retain Set and Test set, this leads to much larger RetDev scores. O and N represent the Original and Retrained models. The Test set MIA against the Original model disinguishes between the Test and Froget sets data points with $73\%$, accuracy while most unlearning methods reach MIA score close to random guessing ($50\%$).}
\label{tab:comp_res_cifar100}
\begin{tabular}{lrrrrrrrrrr}
\toprule
 & RA & FA & TA & RR & FR & TR & RetDev & Indisc & T-MIA & RTE \\
Unlearner &  &  &  &  &  &  &  &  &  &  \\
\midrule
BT & 0.98 & 0.68 & 0.54 & 0.98 & 1.25 & 0.99 & 0.27 & 0.95 & 0.48 & 9.39 \\
CF-k & 1.00 & 0.83 & 0.56 & 1.00 & 1.53 & 1.02 & 0.55 & 0.73 & 0.63 & 5.91 \\
CFW & 0.98 & 0.43 & 0.43 & 0.98 & 0.79 & 0.78 & 0.44 & 1.00 & 0.50 & 6.17 \\
CT & 0.99 & 0.53 & 0.53 & 0.99 & 0.97 & 0.97 & 0.07 & 0.99 & 0.49 & 11.82 \\
EU-k & - & - & - & - & - & - & - & - & - & - \\
FCS & 0.98 & 0.54 & 0.55 & 0.98 & 0.99 & 1.01 & 0.04 & 0.92 & 0.54 & 3.02 \\
FF & - & - & - & - & - & - & - & - & - & - \\
FT & 0.98 & 0.55 & 0.54 & 0.98 & 1.02 & 0.98 & 0.05 & 0.99 & 0.50 & 5.16 \\
GA & 0.34 & 0.33 & 0.24 & 0.34 & 0.60 & 0.44 & 1.61 & 0.90 & 0.55 & 39.97 \\
IU & - & - & - & - & - & - & - & - & - & - \\
KDE & 0.99 & 0.52 & 0.51 & 0.99 & 0.95 & 0.94 & 0.11 & 0.99 & 0.50 & 3.98 \\
MSG & 0.91 & 0.38 & 0.38 & 0.91 & 0.69 & 0.69 & 0.71 & 1.00 & 0.50 & 4.49 \\
NG+ & 0.89 & 0.59 & 0.49 & 0.89 & 1.08 & 0.89 & 0.29 & 0.98 & 0.49 & 12.14 \\
O & 0.98 & 0.98 & 0.56 & 0.98 & 1.81 & 1.02 & 0.85 & 0.53 & 0.73 & 1.10 \\
PRMQ & 0.97 & 0.47 & 0.46 & 0.97 & 0.86 & 0.85 & 0.32 & 1.00 & 0.50 & 4.34 \\
R & 1.00 & 0.55 & 0.55 & 1.00 & 1.00 & 1.00 & 0.00 & 0.99 & 0.49 & 1.00 \\
RNI & 0.99 & 0.45 & 0.45 & 0.99 & 0.83 & 0.82 & 0.36 & 0.98 & 0.49 & 3.65 \\
SCRUB & 0.97 & 0.50 & 0.53 & 0.97 & 0.91 & 0.96 & 0.15 & 0.98 & 0.51 & 3.81 \\
SRL & 1.00 & 0.55 & 0.52 & 1.00 & 1.00 & 0.95 & 0.06 & 0.98 & 0.49 & 3.67 \\
SalUN & 0.98 & 0.49 & 0.51 & 0.98 & 0.91 & 0.93 & 0.18 & 0.99 & 0.49 & 10.66 \\
\bottomrule
\end{tabular}
\end{center}
\end{table*}

\begin{table*}
    \caption{Group Distances: Average scores per performance groups (G1: Best performers, G2: Average performers, G3: Worst performers). This allows for a finer-grained comparison of the groups. For instance, on CIFAR-100, with a \textsc{ResNet18} architecture, and considering the RetDev, the average of the methods from the G3 group is $1.18$ away from the average RetDev of the G2 methods, which is $0.33$ away from G1. Namely, The methods from G3 are $3.58\times$ further apart from the G2 group than G2 from G1}
     \label{table:group_distance}
    \begin{center}
        \begin{tabular}{llccccc}
            \toprule
            \multirow{11}{*}{FashionMNIST} & \multicolumn{6}{c}{Performance Retention Deviation (RetDev)} \\
            \cmidrule{2-7}
            & Architecture & Worst & G3 Mean & G2 Mean & G1 Mean & Best \\
            \cmidrule{2-7}
            & \textsc{ResNet18} & 0.09 & 0.08 & 0.05 & 0.02 & 0.01 \\
            & \textsc{TinyViT} & 0.17 & 0.17 & 0.08 & 0.04 & 0.02 \\
            \cmidrule{2-7}
            & \multicolumn{6}{c}{Indiscernibility} \\ 
            \cmidrule{2-7}
            & Architecture & Worst & G3 Mean & G2 Mean & G1 Mean & Best \\
            \cmidrule{2-7}
            & \textsc{ResNet18} & 0.87 & 0.87 & 0.92 & 0.98 & 0.99 \\
            & \textsc{TinyViT} & 0.93 & 0.93 & 0.98 & 0.99 & 1.00 \\
            \midrule
            \midrule
            \multirow{11}{*}{CIFAR-100} & \multicolumn{6}{c}{Performance Retention Deviation (RetDev)} \\
            \cmidrule{2-7}
            & Architecture & Worst & G3 Mean & G2 Mean & G1 Mean & Best \\
            \cmidrule{2-7}
            & \textsc{ResNet18} & 1.61 & 1.61 & 0.42 & 0.09 & 0.04 \\
            & \textsc{TinyViT} & 2.15 & 2.15 & 0.44 & 0.13 & 0.10 \\
            \cmidrule{2-7}
            & \multicolumn{6}{c}{Indiscernibility} \\ 
            \cmidrule{2-7}
            & Architecture & Worst & G3 Mean & G2 Mean & G1 Mean & Best \\
            \cmidrule{2-7}
            & \textsc{ResNet18} & 0.73 & 0.73 & 0.92 & 0.99 & 1.00 \\
            & \textsc{TinyViT} & 0.87 & 0.88 & 0.94 & 0.99 & 1.00 \\
            \bottomrule
        \end{tabular}
    \end{center}
\end{table*}

\subsection{On the reliability of baselines.}
The commonly used baseline FT trains the original model only on the retain set for several epochs to enable the model to forget information about the forget set. In our evaluation, FT performs best based on the Retention Deviation and is ranked third based on Indiscernibility (Table \ref{table:overall_ranking}). The latter observation may come from the fact that FT does not explicitly unlearn the forget set or perturb the model parameters. Based on these results, we conclude that it is a reasonable baseline against which to evaluate (Tables \ref{app:rank_res}, \ref{app:rank_vit}).
We, however, remark that since the mechanism underlying FT (training on the Retain set to maintain performance) is common to many other methods, these methods may inherit its susceptibility to MIA. Another common baseline, GA, which performs gradient ascent on the forget set, performs poorly across both metrics. Its more recent variation, NG+, which uses an additional retain set correction, ranks fifth for both metrics, making it a more suitable baseline. 

\begin{table*}[t]
\centering
\caption{Run Time Efficiency on ResNet for the top performing methods. CT is the fastest on average, and MSG runs up to 5x faster than naive retraining.} \label{table:rte_resnet18}
\begin{tabular}{lrrrrrr}
\toprule
 & CIFAR-10 & CIFAR-100 & MNIST & FashionMNIST & UTKFace & Average \\
Unlearner &  &  &  &  &  &  \\
\midrule
MSG & 6.80 & 4.49 & 4.29 & 3.32 & 7.57 & 5.29 \\
CFW & 4.67 & 6.17 & 4.29 & 4.90 & 5.54 & 5.11 \\
PRMQ & 4.93 & 4.34 & 3.77 & 3.77 & 5.88 & 4.54 \\
CT & 17.49 & 11.82 & 5.83 & 4.47 & 13.34 & 10.59 \\
KDE & 6.33 & 3.98 & 3.27 & 3.22 & 8.19 & 5.00 \\
FT & 8.15 & 5.16 & 5.07 & 4.29 & 5.78 & 5.69 \\
\bottomrule
\end{tabular}
\end{table*}

\subsection{On the reliability of newly proposed unlearning methods.}
MSG obtained the first rank in both performance retention deviation and indiscernibility (\cref{table:overall_ranking}). Its unique approach identifies the parameters in the Convolutional layers that most contribute to the information being forgotten. This strategy, differing from FT, allows MSG to retain performance from the Retain set while modifying the weights more relevant to the Forget set. PRMQ ranks second in performance retention, making it one of the top performers.
However, it suffers from the same lower performance in Indiscernibility as FT. PRMQ, however, does not leverage the Forget set; instead, it performs a form of knowledge distillation by attempting to reproduce the results of the original models on the Retain set. During the pruning phase, the weights for the MLP and Convolutional layers are reinitialized. CT ranks third in Performance Retention and first in Indiscernibility. It is interesting to note that CT and MSG are consistently among the top performers in Indiscernibility, whereas FT performs poorly (\cref{table:overall_ranking}).

\begin{figure*}[t]
    \centering
    \includegraphics[width=\textwidth]{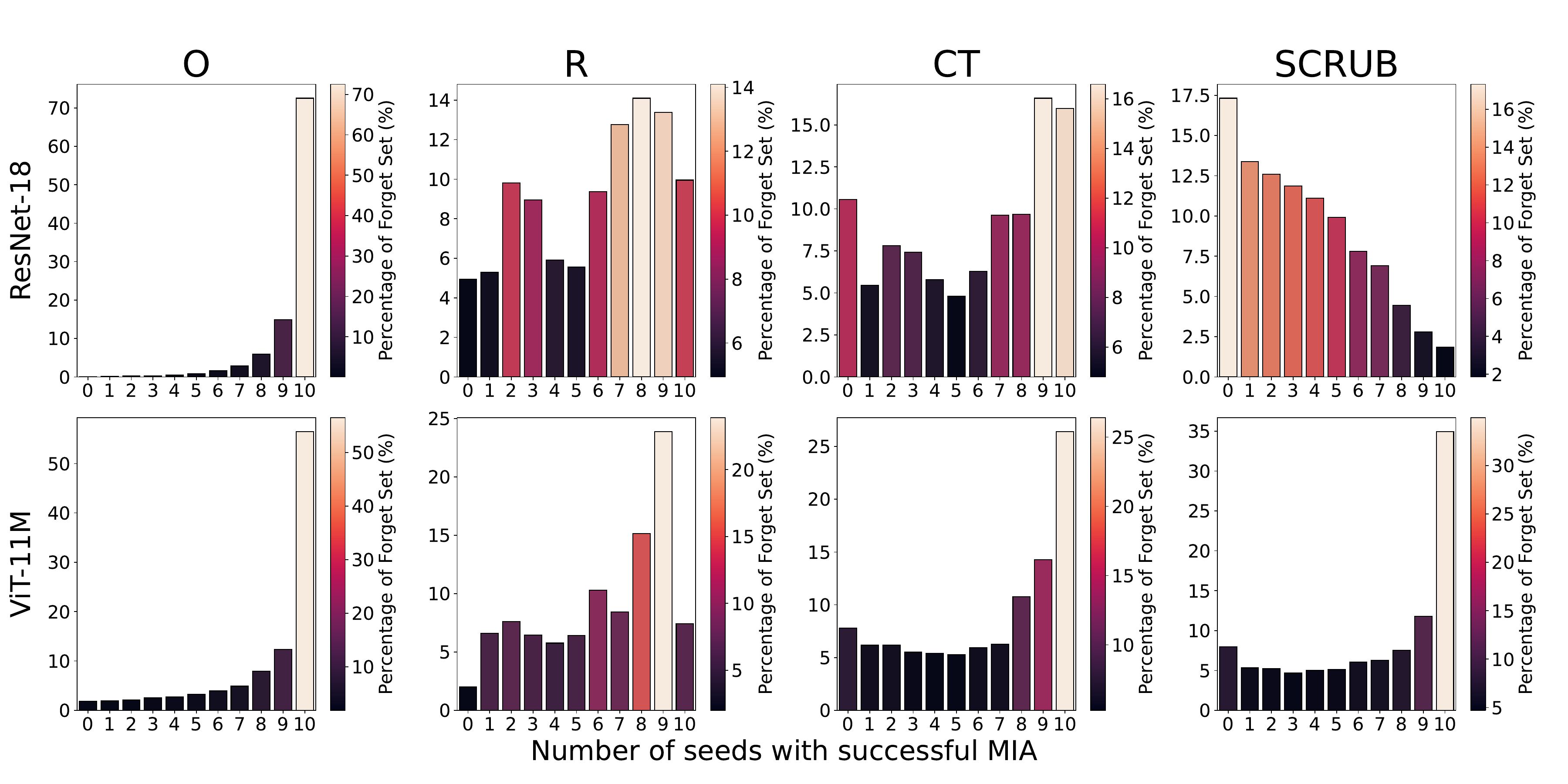}
    \caption{Consistent Unforgettability on CIFAR-100: For each image in the Forget set, we determine the number of models for which the weak MIA is successful for this image. Even for the Retrained model R, $10\%$ of the Forget Set is detected as such across all 10 random model initializations for the ResNet18 model. Nonetheless, for SCRUB close to $17.5\%$, forget sets are never detected as part of the Forget set through the U-MIA. Furthermore, the patterns across architectures differ greatly, especially for SCRUB.}
    \label{figure:consistent_unforget}
\end{figure*}

\subsection{On the robustness across architectures.} 
A critical aspect of a good MU method is its ability to generalize across various DNN architectures. We conducted experiments with both ResNet18 and TinyViT (Tables \ref{app:rank_res}, \cref{app:rank_vit}).
Despite being tailored to Convolutional Neural Networks (CNN) models, methods such as CT still perform competitively when applied to Vision Transformers (\cref{table:overall_ranking}). While proposed for CNN layers, methods such as CT and MSG work well on Vision Transformer as one can leverage the 2D Convolutions used in Positional Encoders.
We provide additional details on the ranking based on architectures in Appendix~\ref{app:architecture_ranking}. 

\subsection{On the speed of the unlearning methods.}
MSG, the best MU candidate, runs ~7.6x faster than retraining from scratch on UTKFace and ~3.3x on FashionMNIST (\cref{table:rte_resnet18}). On average, MSG runs ~5.3x faster than retraining. The fastest method is CT, which achieves a speedup of ~17.5x compared to Retraining on CIFAR-10 and is, on average, $10.59\times$ faster than retraining across all datasets. This speedup stems from its simple approach: transposing the convolutional layer weights.

\color{\mainchange}{
\subsection{On the Consistent Unforgettability.}
We evaluate the concept of Consistent Unforgettability on CIFAR-100 (\cref{figure:consistent_unforget}).
For each method, we count the number of times the membership of a given point of $\mathcal{D}_F$ is correctly inferred when compared to points of the $\mathcal{D}_T$ (\cref{figure:consistent_unforget}).
First, we note that even for the Retrained model, which by design has never seen data points from the Forget set $\mathcal{D}_f$, $10\%$ of the Forget set images are identified as belonging to the Forget set across all $10$ model seeds. Furthermore, $59.6\%$ and $65.2\%$ of the Forget Set images are correctly predicted to belong to the Forget Set for at least five seeds. Furthermore, \cref{figure:consistent_unforget} shows despite using the same hyper-parameters and the same data splits, the performance of an unlearning method can vary across seeds and architecture.
This calls for comparing unlearning models across original model initialization seeds to evaluate the reliability of such a method.
Additionally, we observed data points in the forgot set of CIFAR-100 that are correctly identified across all ten initialization seeds by 13 methods out of 18; the 13 methods might differ from one image to the other. Nonetheless, this shows shared sets of difficult images to forget.
}
\color{black}{}

\begin{figure*}[t]
    \centering
    \includegraphics[width=0.75\textwidth]{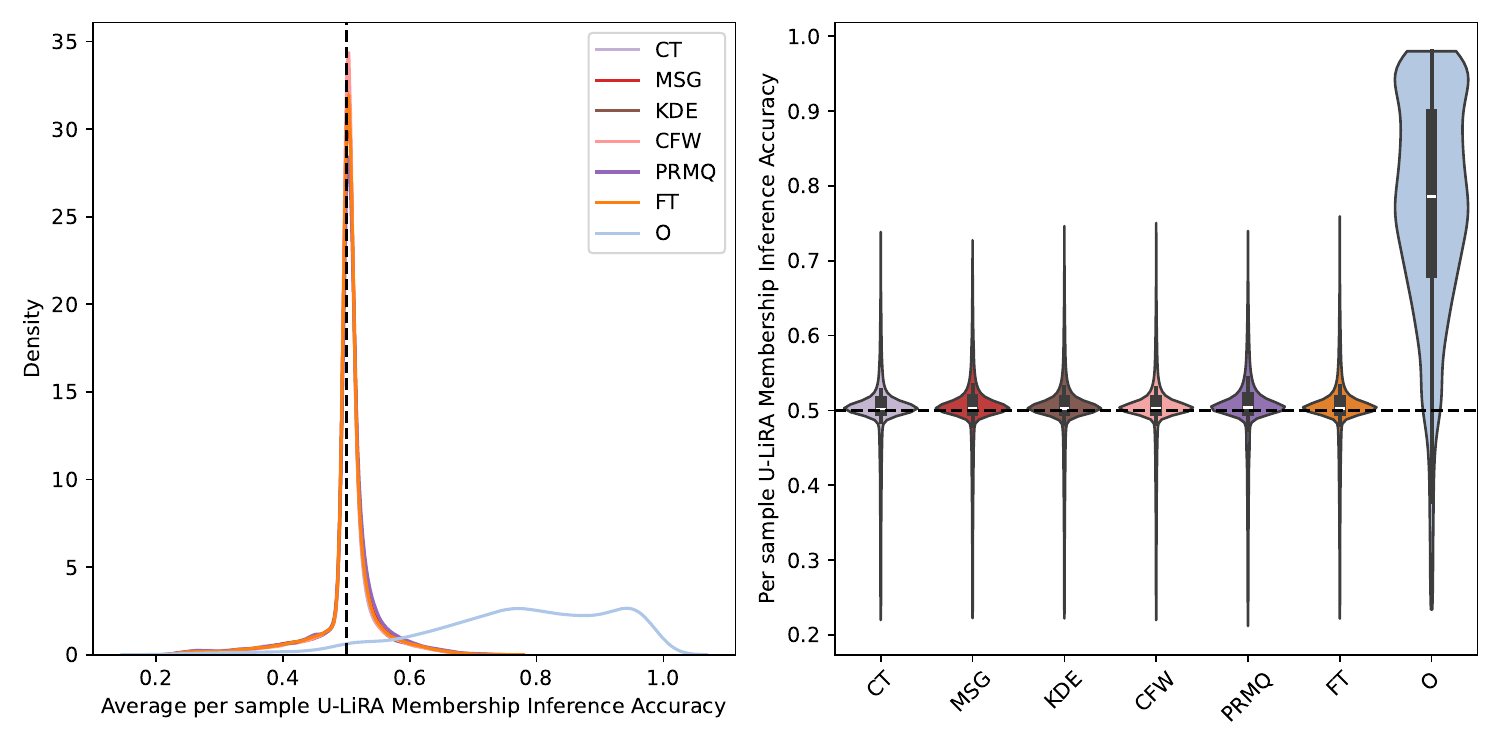}
    \caption{U-LiRA on CIFAR-10 on ResNet models. CT and MSG, which ranked first against U-MIA, showed great resilience against the U-LiRA attack.}
    \label{fig:ulira-perf}
\end{figure*}

\subsection{On the performance when evaluated against a stronger MIA.}
From the results above, we note that MU methods exist that perform fast and reliably across datasets, architectures, and initializations. Specifically, CT, FT, and MSG seem strong candidates for reliable MU methods. However, as has also been highlighted in prior work~\cite{tuReliableEmpiricalMachine2024}, MIA has been debated as a strong metric, as its ability to assess MU is hampered by its ability to infer data membership. In line with this, recent works have introduced more powerful variations on MIA~\cite{yeEnhancedMembershipInference2022}
with~\cite{kurmanjiUnboundedMachineUnlearning2023} proposing the stronger U-LiRA attack for MU.
For the best performers in Table~\ref{table:overall_ranking}, 
we apply the attack setup from~\cite{hayesInexactUnlearningNeeds2024} where we generate a total of 640 models (with varying train, retain, and forget sets) and for each unlearn method, perform a hyperparameter sweep to find the best configuration (for details, see Appendix \ref{app:mias}). 
We determine for each data point its U-LiRA Inference Accuracy and report for each method its average and standard deviation (see Figure \ref{fig:ulira-perf}). From this, we conclude that MSG, as well as CT, resist U-LiRA attacks. Both MSG and CT thus not only rank first in terms of Performance Retention and Indiscernibility based on U-MIA, but also are robust against a stronger variation of MIA. 

\section{Discussion and Conclusion}
\subsection{Conclusion}
The increasing focus on data privacy and the trustworthiness of machine learning models underscores the need for robust and practical methods to unlearn and remove the influence of specific data from trained models.
Due to the growing size of models, we require methods that avoid computationally costly retraining from scratch.
In this work, we comprehensively compared approximate unlearning methods across various models and datasets to address this critical issue.

We compared 18 methods across different datasets and architectures, focusing on assessing the method's ability to maintain privacy and accuracy while being computationally efficient and reliable across datasets, architectures, and random seeds.
Our findings indicate that Masked-Small-Gradients, which accumulates gradients via gradient descent on the data to remember and gradient ascent on the data to forget to determine which weights to update, consistently outperforms all metrics across the datasets, architectures, and initialization seeds.
Similarly, Convolution Transpose, which leverages the simple transposition in convolutional layers, performs well.

CT and MSG were resistant against a population-based Membership Inference Attack (MIA) and a stronger, per-sample attack (U-LiRA). However, a core challenge of approximate unlearning is that these methods will only be as strong as the attacks against which they are tested. As stronger and more complex attacks emerge, some approximate unlearning methods might no longer be as efficient as initially expected. This highlights the need for continuous evaluation and adaptation of unlearning methods to maintain effectiveness. {We also conducted experiments based on L2 distances but found that no method consistently got close to the reference models' weights (see Appendix \ref{app:distance})}.

\subsection{Limitations}
Due to computational costs, we limited our analysis to Tiny Vision Transformers and ResNet; further investigating other architectures could provide useful insights.
We did not investigate different amounts of unlearning samples, which some methods are known to be sensitive to~\cite{kurmanjiUnboundedMachineUnlearning2023}.
We did not consider repeated deletion. Instead, we assume that there is a single forget set and that the unlearning process happens once, as is common in the literature. Nonetheless, in practical applications, one might need to unlearn different smaller forget sets over time, and some unlearning methods might not work as well under such a scenario.
We finally remark once again on the difficulty of evaluation for approximate unlearning~\cite{hayesInexactUnlearningNeeds2024}: while these methods provide significant gains in efficiency, novel attacks might highlight yet unknown weaknesses of the unlearning processes.
This paper focuses exclusively on vision data modalities using vision architectures for image classification. While computer vision is a significant application area, exploring other modalities and architectures to generalize our findings is important.
Different data types and models may rely on distinct inductive biases, and extrapolating our results to these other domains presents challenges.
For example, in audio and sequential data, the interdependence and contextual relationships between data points introduce additional complexities, making unlearning more difficult.
Additionally, this study was conducted with a fixed unlearning budget of 10\% of the data.
While it would be valuable to analyze other unlearning budgets, we prioritized thoroughly investigating this particular budget, which is already commonly used in the community.
Our analysis also focused on unlearning randomly selected data points rather than specific classes, features, or particularly challenging data.
As a result, our findings may not fully generalize to these other scenarios.
Nevertheless, we believe this setting can effectively represent actual deletion requests from users of an image classification service provider.

\subsection{Future work}
First, we focus on natural image data; however, machine unlearning is relevant to other data types, such as medical images or other modalities, such as time series, audio, and speech, or language data.
Second, we focus on the classification task; however, machine unlearning would greatly benefit other learning tasks as well.
For instance removing concepts from generative models for images~\cite{fanSalUnEmpoweringMachine2024} or poisoned data in language models~\cite{hayesInexactUnlearningNeeds2024}.
Third, this work focuses on empirically benchmarking approximate machine unlearning methods.
We do not provide a theoretical analysis of these methods or a rigorous comparison with exact unlearning algorithms.

\subsection{Impact statement} 
This paper aims to highlight the importance of effectively assessing approximate machine unlearning methods.
Our goal is to stress the need to evaluate new unlearning methods against more reliable baselines and experimental setups.
Additionally, it is crucial to assess the consistency of a new unlearning method across various datasets, model architectures, and model initializations.
Without such a thorough evaluation, proposed unlearning methods may provide a false sense of privacy and safety, ultimately limiting their effectiveness for data regulation.

\section*{Acknowledgment}
Xavier F. Cadet is supported by UK Research and Innovation (UKRI Centre for Doctoral Training in AI for Healthcare grant number EP/S023283/1).
Hamed Haddadi is supported by the EPSRC Open Plus Fellowship (EP/W005271/1: Securing the Next Billion Consumer Devices on the Edge).
For open access, the authors have applied a Creative Commons Attribution (CC BY) license to any Author Accepted Manuscript version arising.

\bibliographystyle{IEEEtran}
\bibliography{references}

\clearpage

\begin{appendices}
    \section*{Appendix}
    \addcontentsline{toc}{section}{Appendix}
    The appendix is structured as follows:
    \begin{itemize}
        \item {\bf Per Dataset Results (Section \ref{app:performance})} shows some experimental results in terms of accuracy, retention, privacy metrics, and runtime efficiency per dataset for the 9 combinations of datasets and DNN architectures considered in our work.
        \item {\bf Per Architectures Ranking (Section \ref{app:architecture_ranking})} provides Performance Retention Deviation and Indiscernibility Rankings separated for ResNet18 and TinyViT.
        \item {\bf L2 Distances between Model Weights (Section \ref{app:distance})} shows L2 distances computed between the Unlearned, Original, and Retrained models.
        \item {\bf Data Availability (Section \ref{app:data_availability})} describes the data availability.
        \item {\bf Requirements (Section \ref{app:requirements})} which describes the compute resources.
    \end{itemize}
    \begin{table*}[ht]
\centering
\caption{Summary of the number epochs, learning rate, and batch size for each dataset and model used to train the Original and Retrained models.}
\begin{tabular}{ccccc}
\hline
\textbf{Dataset} & \textbf{Model} & \textbf{Epochs} & \textbf{Learning Rate} & \textbf{Batch Size} \\ \hline
\multirow{2}{*}{FashionMNIST} & \textsc{ResNet18} & 50 & 0.1 & 256 \\ \cline{2-5} 
                                & \textsc{TinyViT} & 50 & 0.1 & 256 \\ \hline
\multirow{2}{*}{MNIST}          & \textsc{ResNet18} & 50 & 0.1 & 256 \\ \cline{2-5} 
                                & \textsc{TinyViT} & 50 & 0.1 & 256 \\ \hline
\multirow{3}{*}{CIFAR-10}        & \textsc{ResNet18} & 182 & 0.1 & 256 \\ \cline{2-5} 
                                & (LiRA) \textsc{ResNet18} & 91 & 0.1 & 256 \\ \cline{2-5} 
                                & \textsc{TinyViT} & 182 & 0.1 & 256 \\ \hline
\multirow{2}{*}{CIFAR-100}       & \textsc{ResNet18} & 182 & 0.1 & 256 \\ \cline{2-5} 
                                & \textsc{TinyViT} & 182 & 0.1 & 256 \\ \hline
\multirow{2}{*}{UTKFace}        & \textsc{ResNet18} & 50 & 0.1 & 128 \\ \cline{2-5} 
                                & \textsc{TinyViT} & 50 & 0.1 & 128 \\ \hline
\end{tabular}
\label{tab:summary}
\end{table*}

\section{Per dataset results} \label{app:performance}

Here, we present the ResNet18 and the TinyViT results across datasets.

{\bf ResNet18}
We provide the tables with Retain Accuracy (RA), Forget Accuracy (FA), Test Accuracy (TA), Retain Retention (RR), Forget Retention (FR), Test Retention (TR), Performance Retention Deviation (RetDev), Indiscinerbility concerning the Test Set (Indisc), U-MIA on the Test set (T-MIA) and RunTime Efficiency (RTE) for every dataset using the ResNet18 model on MNIST (Table \ref{tab:comp_res_mnist}), FashionMNIST (Table \ref{tab:comp_res_fashion}), CIFAR-10 (Table \ref{tab:comp_res_cifar10}), CIFAR-100 (Table \ref{tab:comp_res_cifar100}) and UTKFace (Table \ref{tab:comp_res_utkface}).
In general, CIFAR-100 provides the most visible differences, as the performance on the retain set is much higher than on the test.
Datasets such as MNIST and FashionMNIST tend to show smaller differences between the methods as the performance on both the Retain and Test sets are similar, to begin with.
\begin{table*}
\caption{MNIST - ResNet18}
\label{tab:comp_res_mnist}
\begin{center}
\begin{tabular}{lrrrrrrrrrr}
\toprule
 & RA & FA & TA & RR & FR & TR & RetDev & Indisc & T-MIA & RTE \\
unlearner &  &  &  &  &  &  &  &  &  &  \\
\midrule
BT & 1.00 & 1.00 & 0.99 & 1.00 & 1.01 & 1.00 & 0.01 & 0.99 & 0.50 & 9.76 \\
CF-k & 1.00 & 1.00 & 0.99 & 1.00 & 1.01 & 1.00 & 0.01 & 0.98 & 0.51 & 5.86 \\
CFW & 1.00 & 1.00 & 0.99 & 1.00 & 1.00 & 1.00 & 0.00 & 1.00 & 0.50 & 4.29 \\
CT & 1.00 & 0.99 & 0.99 & 1.00 & 1.00 & 1.00 & 0.00 & 1.00 & 0.50 & 5.83 \\
EU-k & - & - & - & - & - & - & - & - & - & - \\
FCS & 1.00 & 0.99 & 0.99 & 1.00 & 1.00 & 1.00 & 0.00 & 1.00 & 0.50 & 2.11 \\
FF & - & - & - & - & - & - & - & - & - & - \\
FT & 1.00 & 0.99 & 0.99 & 1.00 & 1.00 & 1.00 & 0.00 & 0.99 & 0.51 & 5.07 \\
GA & 0.98 & 0.98 & 0.97 & 0.98 & 0.99 & 0.98 & 0.04 & 0.99 & 0.51 & 33.97 \\
IU & - & - & - & - & - & - & - & - & - & - \\
KDE & 1.00 & 0.99 & 0.99 & 1.00 & 1.00 & 1.00 & 0.00 & 1.00 & 0.50 & 3.27 \\
MSG & 1.00 & 0.99 & 0.99 & 1.00 & 1.00 & 1.00 & 0.00 & 1.00 & 0.50 & 4.29 \\
NG+ & 1.00 & 0.99 & 0.99 & 1.00 & 1.00 & 1.00 & 0.01 & 1.00 & 0.50 & 3.12 \\
O & 1.00 & 1.00 & 0.99 & 1.00 & 1.01 & 1.00 & 0.01 & 0.98 & 0.51 & 1.10 \\
PRMQ & 1.00 & 0.99 & 0.99 & 1.00 & 1.00 & 1.00 & 0.00 & 1.00 & 0.50 & 3.77 \\
R & 1.00 & 0.99 & 0.99 & 1.00 & 1.00 & 1.00 & 0.00 & 1.00 & 0.50 & 1.00 \\
RNI & 1.00 & 1.00 & 1.00 & 1.00 & 1.00 & 1.00 & 0.01 & 1.00 & 0.50 & 4.70 \\
SCRUB & - & - & - & - & - & - & - & - & - & - \\
SRL & 1.00 & 1.00 & 0.99 & 1.00 & 1.01 & 0.99 & 0.01 & 0.99 & 0.51 & 8.55 \\
SalUN & 1.00 & 1.00 & 0.99 & 1.00 & 1.00 & 0.99 & 0.01 & 0.99 & 0.50 & 3.19 \\
\bottomrule
\end{tabular}
\end{center}
\end{table*}

\begin{table*}
\caption{FashionMNIST - ResNet18}
\label{tab:comp_res_fashion}
\begin{center}
\begin{tabular}{lrrrrrrrrrr}
\toprule
 & RA & FA & TA & RR & FR & TR & RetDev & Indisc & T-MIA & RTE \\
Unlearner &  &  &  &  &  &  &  &  &  &  \\
\midrule
BT & 1.00 & 0.96 & 0.92 & 1.00 & 1.03 & 0.99 & 0.04 & 0.98 & 0.51 & 13.57 \\
CF-k & 0.98 & 0.97 & 0.91 & 0.98 & 1.05 & 0.98 & 0.09 & 0.92 & 0.54 & 16.31 \\
CFW & 1.00 & 0.95 & 0.92 & 1.00 & 1.02 & 0.99 & 0.03 & 0.97 & 0.51 & 4.90 \\
CT & 1.00 & 0.92 & 0.92 & 1.00 & 0.99 & 0.99 & 0.02 & 0.99 & 0.50 & 4.47 \\
EU-k & - & - & - & - & - & - & - & - & - & - \\
FCS & 0.98 & 0.93 & 0.91 & 0.98 & 1.00 & 0.98 & 0.04 & 0.98 & 0.51 & 2.79 \\
FF & - & - & - & - & - & - & - & - & - & - \\
FT & 1.00 & 0.95 & 0.92 & 1.00 & 1.02 & 1.00 & 0.02 & 0.98 & 0.51 & 4.29 \\
GA & - & - & - & - & - & - & - & - & - & - \\
IU & 1.00 & 1.00 & 0.93 & 1.00 & 1.08 & 1.00 & 0.08 & 0.87 & 0.56 & 14.97 \\
KDE & 1.00 & 0.93 & 0.92 & 1.00 & 1.00 & 1.00 & 0.01 & 0.99 & 0.50 & 3.22 \\
MSG & 1.00 & 0.93 & 0.91 & 1.00 & 1.00 & 0.99 & 0.01 & 0.98 & 0.51 & 3.32 \\
NG+ & 0.99 & 0.94 & 0.91 & 0.99 & 1.01 & 0.99 & 0.03 & 0.99 & 0.49 & 3.21 \\
O & 1.00 & 1.00 & 0.93 & 1.00 & 1.08 & 1.00 & 0.08 & 0.87 & 0.56 & 1.11 \\
PRMQ & 0.98 & 0.93 & 0.91 & 0.98 & 1.00 & 0.99 & 0.04 & 0.98 & 0.51 & 3.77 \\
R & 1.00 & 0.93 & 0.92 & 1.00 & 1.00 & 1.00 & 0.00 & 1.00 & 0.50 & 1.00 \\
RNI & 0.98 & 0.93 & 0.91 & 0.98 & 1.00 & 0.98 & 0.04 & 0.98 & 0.51 & 3.25 \\
SCRUB & 0.95 & 0.93 & 0.90 & 0.95 & 1.00 & 0.98 & 0.08 & 0.97 & 0.51 & 5.62 \\
SRL & 1.00 & 0.97 & 0.92 & 1.00 & 1.04 & 1.00 & 0.05 & 0.98 & 0.51 & 24.85 \\
SalUN & 0.99 & 0.97 & 0.92 & 0.99 & 1.04 & 0.99 & 0.06 & 0.98 & 0.51 & 25.02 \\
\bottomrule
\end{tabular}
\end{center}
\end{table*}

\begin{table*}
\caption{CIFAR-10 - ResNet18}
\label{tab:comp_res_cifar10}
\begin{center}
\begin{tabular}{lrrrrrrrrrr}
\toprule
 & RA & FA & TA & RR & FR & TR & RetDev & Indisc & T-MIA & RTE \\
Unlearner &  &  &  &  &  &  &  &  &  &  \\
\midrule
BT & 0.94 & 0.87 & 0.84 & 0.94 & 1.00 & 0.96 & 0.10 & 0.97 & 0.48 & 51.96 \\
CF-k & - & - & - & - & - & - & - & - & - & -\\
CFW & 1.00 & 0.81 & 0.80 & 1.00 & 0.92 & 0.92 & 0.16 & 1.00 & 0.50 & 4.67 \\
CT & 1.00 & 0.82 & 0.81 & 1.00 & 0.93 & 0.93 & 0.14 & 0.99 & 0.50 & 17.49 \\
EU-k & - & - & - & - & - & - & - & - & - & - \\
FCS & 0.99 & 0.86 & 0.84 & 0.99 & 0.98 & 0.96 & 0.07 & 0.98 & 0.49 & 22.53 \\
FF & - & - & - & - & - & - & - & - & - & - \\
FT & 1.00 & 0.84 & 0.82 & 1.00 & 0.96 & 0.95 & 0.09 & 1.00 & 0.50 & 8.15 \\
GA & 0.91 & 0.89 & 0.81 & 0.91 & 1.02 & 0.93 & 0.18 & 0.92 & 0.54 & 91.48 \\
IU & 0.95 & 0.94 & 0.84 & 0.95 & 1.08 & 0.97 & 0.16 & 0.91 & 0.55 & 64.61 \\
KDE & 0.98 & 0.84 & 0.80 & 0.98 & 0.96 & 0.92 & 0.15 & 0.97 & 0.52 & 6.33 \\
MSG & 1.00 & 0.85 & 0.83 & 1.00 & 0.97 & 0.95 & 0.08 & 0.99 & 0.51 & 6.80 \\
NG+ & 0.97 & 0.89 & 0.85 & 0.98 & 1.02 & 0.97 & 0.07 & 0.98 & 0.51 & 12.89 \\
O & 0.96 & 0.96 & 0.85 & 0.96 & 1.10 & 0.98 & 0.16 & 0.89 & 0.55 & 1.08 \\
PRMQ & 1.00 & 0.86 & 0.83 & 1.00 & 0.98 & 0.95 & 0.07 & 0.98 & 0.51 & 4.93 \\
R & 1.00 & 0.87 & 0.87 & 1.00 & 1.00 & 1.00 & 0.00 & 1.00 & 0.50 & 1.00 \\
RNI & 1.00 & 0.83 & 0.81 & 1.00 & 0.95 & 0.93 & 0.12 & 0.99 & 0.50 & 3.60 \\
SCRUB & 0.99 & 0.85 & 0.85 & 0.99 & 0.97 & 0.98 & 0.07 & 0.99 & 0.50 & 2.57 \\
SRL & 0.99 & 0.93 & 0.84 & 0.99 & 1.06 & 0.97 & 0.10 & 0.98 & 0.49 & 5.52 \\
SalUN & 0.98 & 0.90 & 0.84 & 0.98 & 1.04 & 0.97 & 0.08 & 0.96 & 0.48 & 18.04 \\
\bottomrule
\end{tabular}
\end{center}
\end{table*}

\begin{table*}
\caption{UTKFace - ResNet18}
\label{tab:comp_res_utkface}
\begin{center}
\begin{tabular}{lrrrrrrrrrr}
\toprule
 & RA & FA & TA & RR & FR & TR & RetDev & Indisc & T-MIA & RTE \\
Unlearner &  &  &  &  &  &  &  &  &  &  \\
\midrule
BT & 1.00 & 0.74 & 0.73 & 1.00 & 1.00 & 0.96 & 0.04 & 0.99 & 0.50 & 12.48 \\
CF-k & 1.00 & 1.00 & 0.75 & 1.00 & 1.34 & 0.99 & 0.35 & 0.70 & 0.65 & 5.35 \\
CFW & 1.00 & 0.76 & 0.76 & 1.00 & 1.02 & 1.00 & 0.02 & 1.00 & 0.50 & 5.54 \\
CT & 1.00 & 0.75 & 0.76 & 1.00 & 1.01 & 1.00 & 0.01 & 0.99 & 0.50 & 13.34 \\
EU-k & 0.72 & 0.61 & 0.59 & 0.72 & 0.82 & 0.77 & 0.68 & 0.99 & 0.51 & 11.42 \\
FCS & 0.90 & 0.70 & 0.70 & 0.91 & 0.94 & 0.93 & 0.23 & 0.99 & 0.50 & 4.33 \\
FF & - & - & - & - & - & - & - & - & - & - \\
FT & 1.00 & 0.76 & 0.77 & 1.00 & 1.02 & 1.01 & 0.04 & 1.00 & 0.50 & 5.78 \\
GA & 0.49 & 0.47 & 0.40 & 0.49 & 0.63 & 0.53 & 1.34 & 0.92 & 0.54 & 235.10 \\
IU & 1.00 & 1.00 & 0.76 & 1.00 & 1.34 & 1.01 & 0.35 & 0.62 & 0.69 & 33.77 \\
KDE & 0.99 & 0.79 & 0.76 & 0.99 & 1.06 & 1.00 & 0.07 & 0.97 & 0.52 & 8.19 \\
MSG & 1.00 & 0.80 & 0.76 & 1.00 & 1.08 & 1.00 & 0.08 & 0.96 & 0.52 & 7.57 \\
NG+ & 0.94 & 0.80 & 0.72 & 0.95 & 1.07 & 0.95 & 0.18 & 0.99 & 0.51 & 6.73 \\
O & 1.00 & 1.00 & 0.76 & 1.00 & 1.34 & 1.01 & 0.35 & 0.61 & 0.69 & 1.09 \\
PRMQ & 0.91 & 0.72 & 0.72 & 0.91 & 0.97 & 0.95 & 0.17 & 1.00 & 0.50 & 5.88 \\
R & 1.00 & 0.75 & 0.76 & 1.00 & 1.00 & 1.00 & 0.00 & 1.00 & 0.50 & 1.00 \\
RNI & 0.96 & 0.75 & 0.73 & 0.96 & 1.01 & 0.96 & 0.08 & 0.98 & 0.51 & 5.11 \\
SCRUB & 0.80 & 0.76 & 0.69 & 0.80 & 1.01 & 0.92 & 0.29 & 0.94 & 0.53 & 4.64 \\
SRL & 1.00 & 0.80 & 0.73 & 1.00 & 1.08 & 0.97 & 0.11 & 0.99 & 0.51 & 12.05 \\
SalUN & 0.97 & 0.79 & 0.73 & 0.98 & 1.06 & 0.96 & 0.12 & 0.97 & 0.52 & 36.80 \\
\bottomrule
\end{tabular}
\end{center}
\end{table*}

{\bf TinyViT}
We provide the tables with RA, FA, TA, RR, FR, TR, RetDev, Indisc, T-MIA and RTE for MNIST (Table \ref{tab:comp_vit_mnist}), FashionMNIST (Table \ref{tab:comp_vit_fashion}), CIFAR-10 (Table \ref{tab:comp_vit_cifar10}) and CIFAR-100 (Table \ref{tab:comp_vit_cifar100}) using the TinyViT model.

\begin{table*}
\centering
\caption{MNIST - TinyViT}
\label{tab:comp_vit_mnist}
\begin{tabular}{lrrrrrrrrrr}
\toprule
 & RA & FA & TA & RR & FR & TR & RetDev & Indisc & T-MIA & RTE \\
Unlearner &  &  &  &  &  &  &  &  &  &  \\
\midrule
BT & 1.00 & 1.00 & 0.99 & 1.00 & 1.01 & 1.00 & 0.01 & 1.00 & 0.50 & 4.39 \\
CF-k & 1.00 & 1.00 & 0.99 & 1.00 & 1.01 & 1.00 & 0.01 & 0.99 & 0.51 & 123.88 \\
CFW & 1.00 & 0.99 & 0.99 & 1.00 & 1.00 & 1.00 & 0.00 & 0.99 & 0.50 & 5.15 \\
CT & 1.00 & 0.99 & 0.99 & 1.00 & 1.00 & 1.00 & 0.00 & 1.00 & 0.50 & 5.92 \\
EU-k & 1.00 & 1.00 & 0.99 & 1.00 & 1.01 & 1.00 & 0.01 & 0.99 & 0.50 & 11.93 \\
FCS & 1.00 & 1.00 & 0.99 & 1.00 & 1.01 & 1.00 & 0.01 & 0.99 & 0.49 & 7.42 \\
FF & - & - & - & - & - & - & - & - & - & - \\
FT & 1.00 & 0.99 & 0.99 & 1.00 & 1.00 & 1.00 & 0.01 & 1.00 & 0.50 & 7.69 \\
GA & 0.97 & 0.97 & 0.96 & 0.97 & 0.98 & 0.97 & 0.08 & 0.99 & 0.51 & 390.99 \\
IU & - & - & - & - & - & - & - & - & - & - \\
KDE & 1.00 & 0.99 & 0.99 & 1.00 & 1.00 & 1.00 & 0.01 & 1.00 & 0.50 & 5.35 \\
MSG & 1.00 & 0.99 & 0.99 & 1.00 & 1.00 & 1.00 & 0.00 & 1.00 & 0.50 & 5.21 \\
NG+ & 1.00 & 0.99 & 0.99 & 1.00 & 1.00 & 1.00 & 0.00 & 0.99 & 0.50 & 2.74 \\
O & 1.00 & 1.00 & 0.99 & 1.00 & 1.01 & 1.00 & 0.01 & 0.99 & 0.51 & 0.97 \\
PRMQ & 1.00 & 0.99 & 0.99 & 1.00 & 1.00 & 1.00 & 0.00 & 1.00 & 0.50 & 6.88 \\
R & 1.00 & 0.99 & 0.99 & 1.00 & 1.00 & 1.00 & 0.00 & 1.00 & 0.50 & 1.00 \\
RNI & 1.00 & 0.99 & 0.99 & 1.00 & 1.00 & 1.00 & 0.01 & 1.00 & 0.50 & 8.03 \\
SCRUB & 0.99 & 0.99 & 0.99 & 0.99 & 1.00 & 1.00 & 0.01 & 1.00 & 0.50 & 3.54 \\
SRL & 1.00 & 0.99 & 0.99 & 1.00 & 1.00 & 1.00 & 0.01 & 0.98 & 0.51 & 16.37 \\
SalUN & 1.00 & 1.00 & 0.99 & 1.00 & 1.01 & 0.99 & 0.01 & 0.99 & 0.50 & 43.67 \\
\bottomrule
\end{tabular}
\end{table*}

\begin{table*}
\centering
\caption{FashionMNIST - TinyViT}
\label{tab:comp_vit_fashion}
\begin{tabular}{lrrrrrrrrrr}
\toprule
 & RA & FA & TA & RR & FR & TR & RetDev & Indisc & T-MIA & RTE \\
Unlearner &  &  &  &  &  &  &  &  &  &  \\
\midrule
BT & 0.97 & 0.94 & 0.91 & 0.97 & 1.01 & 0.99 & 0.04 & 0.98 & 0.51 & 3.48 \\
CF-k & - & - & - & - & - & - & - & - & - & - \\
CFW & 0.99 & 0.94 & 0.92 & 0.99 & 1.01 & 1.00 & 0.02 & 0.99 & 0.51 & 5.40 \\
CT & 0.98 & 0.92 & 0.91 & 0.98 & 0.99 & 0.99 & 0.04 & 0.99 & 0.50 & 6.00 \\
EU-k & 0.95 & 0.94 & 0.91 & 0.95 & 1.01 & 0.99 & 0.07 & 0.97 & 0.51 & 5.34 \\
FCS & 0.98 & 0.93 & 0.91 & 0.98 & 1.01 & 0.99 & 0.04 & 0.98 & 0.51 & 4.58 \\
FF & - & - & - & - & - & - & - & - & - & - \\
FT & 0.99 & 0.94 & 0.92 & 1.00 & 1.01 & 1.00 & 0.02 & 0.98 & 0.51 & 5.12 \\
GA & 0.92 & 0.91 & 0.85 & 0.92 & 0.99 & 0.93 & 0.17 & 0.93 & 0.53 & 50.72 \\
IU & - & - & - & - & - & - & - & - & - & - \\
KDE & 1.00 & 0.94 & 0.92 & 1.00 & 1.02 & 1.00 & 0.02 & 0.98 & 0.51 & 3.38 \\
MSG & 0.96 & 0.92 & 0.91 & 0.96 & 1.00 & 0.99 & 0.05 & 0.99 & 0.51 & 8.21 \\
NG+ & 0.97 & 0.92 & 0.91 & 0.97 & 1.00 & 0.99 & 0.05 & 0.98 & 0.51 & 13.65 \\
O & 1.00 & 1.00 & 0.92 & 1.00 & 1.08 & 1.00 & 0.08 & 0.89 & 0.56 & 0.97 \\
PRMQ & 0.98 & 0.94 & 0.91 & 0.98 & 1.01 & 0.99 & 0.04 & 0.98 & 0.51 & 4.67 \\
R & 1.00 & 0.93 & 0.92 & 1.00 & 1.00 & 1.00 & 0.00 & 1.00 & 0.50 & 1.00 \\
RNI & 0.97 & 0.94 & 0.91 & 0.97 & 1.01 & 0.99 & 0.05 & 0.97 & 0.51 & 5.00 \\
SCRUB & 0.96 & 0.95 & 0.91 & 0.96 & 1.03 & 0.99 & 0.08 & 0.96 & 0.52 & 9.56 \\
SRL & 0.98 & 0.94 & 0.91 & 0.98 & 1.01 & 0.99 & 0.04 & 1.00 & 0.50 & 9.41 \\
SalUN & 0.97 & 0.94 & 0.91 & 0.97 & 1.01 & 0.99 & 0.05 & 0.99 & 0.51 & 6.53 \\
\bottomrule
\end{tabular}
\end{table*}

\begin{table*}
\centering
\caption{CIFAR-10 - TinyViT}
\label{tab:comp_vit_cifar10}
\begin{tabular}{lrrrrrrrrrr}
\toprule
 & RA & FA & TA & RR & FR & TR & RetDev & Indisc & T-MIA & RTE \\
Unlearner &  &  &  &  &  &  &  &  &  &  \\
\midrule
BT & 0.91 & 0.91 & 0.85 & 0.91 & 1.02 & 0.97 & 0.14 & 0.99 & 0.50 & 4.13 \\
CF-k & 0.99 & 0.89 & 0.84 & 0.99 & 1.00 & 0.95 & 0.06 & 0.96 & 0.52 & 5.18 \\
CFW & 0.98 & 0.87 & 0.84 & 0.99 & 0.98 & 0.96 & 0.07 & 0.98 & 0.51 & 28.85 \\
CT & 0.98 & 0.82 & 0.81 & 0.98 & 0.93 & 0.92 & 0.17 & 1.00 & 0.50 & 23.48 \\
EU-k & 0.90 & 0.90 & 0.84 & 0.90 & 1.02 & 0.95 & 0.16 & 0.97 & 0.52 & 43.25 \\
FCS & 0.98 & 0.84 & 0.83 & 0.98 & 0.95 & 0.94 & 0.13 & 0.99 & 0.49 & 5.15 \\
FF & - & - & - & - & - & - & - & - & - & - \\
FT & 1.00 & 0.87 & 0.84 & 1.00 & 0.98 & 0.95 & 0.07 & 0.98 & 0.51 & 5.85 \\
GA & 0.85 & 0.85 & 0.80 & 0.85 & 0.96 & 0.91 & 0.29 & 0.97 & 0.52 & 514.77 \\
IU & - & - & - & - & - & - & - & - & - & - \\
KDE & 0.97 & 0.86 & 0.84 & 0.97 & 0.97 & 0.96 & 0.11 & 0.99 & 0.50 & 5.27 \\
MSG & 1.00 & 0.85 & 0.83 & 1.00 & 0.96 & 0.94 & 0.10 & 0.99 & 0.51 & 7.38 \\
NG+ & 0.93 & 0.86 & 0.85 & 0.93 & 0.97 & 0.96 & 0.14 & 0.99 & 0.50 & 4.10 \\
O & 0.92 & 0.92 & 0.86 & 0.92 & 1.04 & 0.97 & 0.15 & 0.95 & 0.53 & 0.97 \\
PRMQ & 1.00 & 0.87 & 0.84 & 1.00 & 0.98 & 0.95 & 0.07 & 0.99 & 0.51 & 4.00 \\
R & 1.00 & 0.89 & 0.88 & 1.00 & 1.00 & 1.00 & 0.00 & 1.00 & 0.50 & 1.00 \\
RNI & 0.97 & 0.84 & 0.81 & 0.98 & 0.95 & 0.92 & 0.15 & 0.98 & 0.51 & 6.50 \\
SCRUB & 1.00 & 0.84 & 0.84 & 1.00 & 0.95 & 0.95 & 0.10 & 0.99 & 0.50 & - \\
SRL & 0.97 & 0.88 & 0.84 & 0.97 & 0.99 & 0.96 & 0.08 & 0.99 & 0.49 & 8.52 \\
SalUN & 0.96 & 0.89 & 0.85 & 0.96 & 1.00 & 0.96 & 0.08 & 0.99 & 0.50 & 8.30 \\
\bottomrule
\end{tabular}
\end{table*}

\begin{table*}
\centering
\caption{CIFAR-100 - TinyViT}
\label{tab:comp_vit_cifar100}
\begin{tabular}{lrrrrrrrrrr}
\toprule
 & RA & FA & TA & RR & FR & TR & RetDev & Indisc & T-MIA & RTE \\
Unlearner &  &  &  &  &  &  &  &  &  &  \\
\midrule
BT & 0.82 & 0.66 & 0.57 & 0.82 & 1.11 & 0.96 & 0.33 & 0.94 & 0.53 & 31.63 \\
CF-k & 0.24 & 0.18 & 0.18 & 0.24 & 0.31 & 0.30 & 2.15 & 0.98 & 0.51 & 5.97 \\
CFW & 0.98 & 0.58 & 0.56 & 0.98 & 0.97 & 0.95 & 0.10 & 0.99 & 0.51 & 9.18 \\
CT & 0.97 & 0.55 & 0.55 & 0.98 & 0.93 & 0.93 & 0.17 & 0.99 & 0.49 & 9.80 \\
EU-k & 0.61 & 0.60 & 0.49 & 0.61 & 1.01 & 0.81 & 0.59 & 0.90 & 0.55 & 19.29 \\
FCS & 0.89 & 0.60 & 0.58 & 0.89 & 1.01 & 0.98 & 0.13 & 0.97 & 0.48 & 7.05 \\
FF & - & - & - & - & - & - & - & - & - & - \\
FT & 1.00 & 0.56 & 0.55 & 1.00 & 0.94 & 0.91 & 0.15 & 1.00 & 0.50 & 5.90 \\
GA & 0.60 & 0.58 & 0.46 & 0.60 & 0.97 & 0.77 & 0.66 & 0.87 & 0.56 & 91.91 \\
IU & - & - & - & - & - & - & - & - & - & - \\
KDE & 0.93 & 0.58 & 0.57 & 0.94 & 0.98 & 0.96 & 0.13 & 0.99 & 0.50 & 4.03 \\
MSG & 0.97 & 0.57 & 0.56 & 0.97 & 0.95 & 0.93 & 0.15 & 1.00 & 0.50 & 5.89 \\
NG+ & 0.84 & 0.57 & 0.55 & 0.84 & 0.95 & 0.92 & 0.29 & 0.96 & 0.52 & 3.01 \\
O & 0.87 & 0.87 & 0.61 & 0.87 & 1.46 & 1.02 & 0.61 & 0.74 & 0.63 & 0.98 \\
PRMQ & 0.95 & 0.62 & 0.57 & 0.95 & 1.03 & 0.96 & 0.13 & 0.95 & 0.52 & 5.58 \\
R & 1.00 & 0.60 & 0.60 & 1.00 & 1.00 & 1.00 & 0.00 & 1.00 & 0.50 & 1.00 \\
RNI & 0.85 & 0.52 & 0.51 & 0.85 & 0.87 & 0.85 & 0.42 & 0.99 & 0.50 & 5.38 \\
SCRUB & 0.77 & 0.64 & 0.57 & 0.77 & 1.08 & 0.96 & 0.35 & 0.93 & 0.53 & 6.17 \\
SRL & 0.98 & 0.57 & 0.57 & 0.98 & 0.96 & 0.96 & 0.11 & 0.97 & 0.49 & 5.92 \\
SalUN & 0.97 & 0.58 & 0.57 & 0.97 & 0.97 & 0.96 & 0.10 & 0.98 & 0.49 & 7.03 \\
\bottomrule
\end{tabular}
\end{table*}

\section{Per architectures rankings} \label{app:architecture_ranking}
Here, we present the rankings across datasets for ResNet18 (Table \ref{app:rank_res}) and TinyVit (Table \ref{app:rank_vit}).
We note that some methods, such as RNI or NG+, are less efficient regarding incernibility on the ViT architectures. However, methods such as SCRUB are less efficient regarding Retention Deviation on the ViT architecture.

\begin{table*}[ht]
\centering
\caption{Ranking on ResNet}
\label{app:rank_res}
\begin{tabular}{llcccccllcccc}
\toprule
& & \multicolumn{4}{c}{Retention Deviation} & & & & \multicolumn{4}{c}{Indiscernibility} \\
\cmidrule(r){3-6} \cmidrule(r){10-13}
Rank & Method & G1 & G2 & G3 & F & & Rank & Method & G1 & G2 & G3 & F \\
\midrule
1 & FT & 5 & 0 & 0 & 0 & & 1 & CFW & 5 & 0 & 0 & 0 \\
2 & FCS & 4 & 1 & 0 & 0 & & 1 & CT & 5 & 0 & 0 & 0 \\
2 & MSG & 4 & 1 & 0 & 0 & & 1 & MSG & 5 & 0 & 0 & 0 \\
3 & CT & 4 & 0 & 1 & 0 & & 1 & RNI & 5 & 0 & 0 & 0 \\
3 & KDE & 4 & 0 & 1 & 0 & & 2 & FT & 4 & 1 & 0 & 0 \\
4 & NG+ & 3 & 2 & 0 & 0 & & 2 & KDE & 4 & 1 & 0 & 0 \\
4 & PRMQ & 3 & 2 & 0 & 0 & & 2 & NG+ & 4 & 1 & 0 & 0 \\
4 & SalUN & 3 & 2 & 0 & 0 & & 2 & PRMQ & 4 & 1 & 0 & 0 \\
5 & CFW & 3 & 1 & 1 & 0 & & 3 & FCS & 3 & 2 & 0 & 0 \\
6 & SCRUB & 3 & 0 & 1 & 1 & & 3 & SRL & 3 & 2 & 0 & 0 \\
7 & SRL & 2 & 3 & 0 & 0 & & 3 & SalUN & 3 & 2 & 0 & 0 \\
8 & BT & 1 & 4 & 0 & 0 & & 4 & SCRUB & 3 & 1 & 0 & 1 \\
8 & RNI & 1 & 4 & 0 & 0 & & 5 & BT & 2 & 3 & 0 & 0 \\
9 & CF-k & 1 & 2 & 1 & 1 & & 6 & EU-k & 1 & 0 & 0 & 4 \\
10 & IU & 1 & 0 & 2 & 2 & & 7 & GA & 0 & 3 & 1 & 1 \\
11 & EU-k & 0 & 1 & 0 & 4 & & 8 & CF-k & 0 & 1 & 3 & 1 \\
12 & GA & 0 & 0 & 4 & 1 & & 9 & IU & 0 & 0 & 3 & 2 \\
13 & FF & 0 & 0 & 0 & 5 & & 10 & FF & 0 & 0 & 0 & 5 \\

\bottomrule
\end{tabular}
\end{table*}

\begin{table*}[ht]
\centering
\caption{Ranking of ViT}
\label{app:rank_vit}
\begin{tabular}{llcccccllcccc}
\toprule
& & \multicolumn{4}{c}{Retention Deviation} & & & & \multicolumn{4}{c}{Indiscernibility} \\
\cmidrule(r){3-6} \cmidrule(r){10-13}
Rank & Method & G1 & G2 & G3 & F & & Rank & Method & G1 & G2 & G3 & F \\
\midrule
1 & CFW & 4 & 0 & 0 & 0 & & 1 & CT & 4 & 0 & 0 & 0 \\
1 & MSG & 4 & 0 & 0 & 0 & & 1 & MSG & 4 & 0 & 0 & 0 \\
1 & PRMQ & 4 & 0 & 0 & 0 & & 2 & KDE & 3 & 1 & 0 & 0 \\
2 & CT & 3 & 1 & 0 & 0 & & 2 & SalUN & 3 & 1 & 0 & 0 \\
2 & FT & 3 & 1 & 0 & 0 & & 3 & SRL & 3 & 0 & 1 & 0 \\
2 & KDE & 3 & 1 & 0 & 0 & & 4 & BT & 2 & 2 & 0 & 0 \\
2 & SRL & 3 & 1 & 0 & 0 & & 4 & CFW & 2 & 2 & 0 & 0 \\
2 & SalUN & 3 & 1 & 0 & 0 & & 4 & FCS & 2 & 2 & 0 & 0 \\
3 & FCS & 2 & 2 & 0 & 0 & & 4 & FT & 2 & 2 & 0 & 0 \\
3 & NG+ & 2 & 2 & 0 & 0 & & 4 & PRMQ & 2 & 2 & 0 & 0 \\
4 & BT & 1 & 3 & 0 & 0 & & 4 & RNI & 2 & 2 & 0 & 0 \\
4 & RNI & 1 & 3 & 0 & 0 & & 4 & SCRUB & 2 & 2 & 0 & 0 \\
4 & SCRUB & 1 & 3 & 0 & 0 & & 5 & NG+ & 1 & 3 & 0 & 0 \\
5 & CF-k & 1 & 1 & 1 & 1 & & 6 & CF-k & 1 & 1 & 1 & 1 \\
6 & EU-k & 0 & 4 & 0 & 0 & & 7 & EU-k & 0 & 2 & 2 & 0 \\
7 & GA & 0 & 1 & 3 & 0 & & 8 & GA & 0 & 1 & 3 & 0 \\
8 & FF & 0 & 0 & 0 & 4 & & 9 & FF & 0 & 0 & 0 & 4 \\
8 & IU & 0 & 0 & 0 & 4 & & 9 & IU & 0 & 0 & 0 & 4 \\
\bottomrule
\end{tabular}
\end{table*}

\section{L2 Distances between model weights} \label{app:distance}
\begin{figure*}[ht]
    \centering
    \includegraphics[width=\textwidth]{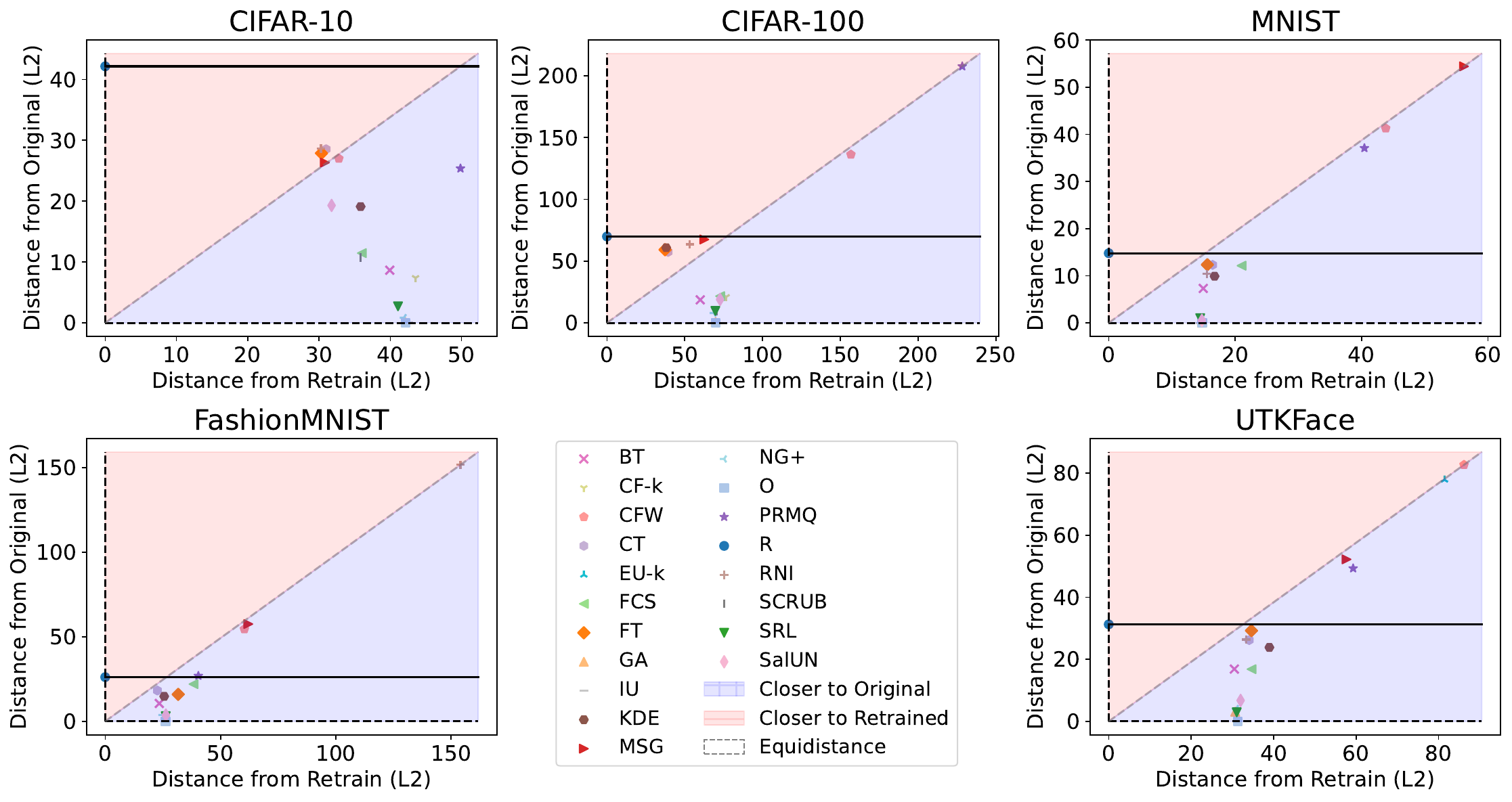}
    \caption{L2 Distance between the Unlearned ResNet18 models, the Original and Retrained models. None of the unlearned models gets close to the Retrained model's weights; most unlearned Models are closer to the Original model than the Retrained model.}
    \label{fig:l2_distances}
\end{figure*}

The distance between the Unlearned and Retrained models has also been considered in the literature to evaluate MU.
Nevertheless, we observe that models end up at a similar distance to the Retrained model, with significant differences in performance.
We further note that one challenging aspect of the L2 distance comparison is the different factors of Weight Decay used by the MU method.
The hyper-parameter searches determine these Weight Decay factors, which can significantly vary from one unlearning method to another, making it challenging to compare methods.
Furthermore, the best-performing method, MSG, is usually at the same distance as the Original and Retained models.
For each method, for each initialization seed, we computed the L2 distance between the unlearned model $f_U$ and the retrained model $f_R$, as well as between the $f_U$ and $f_O$ (Figure \ref{fig:l2_distances}).

Although having the same weight as the Retrained model would indicate that the unlearned model has unlearned $\mathcal{D}_F$, our evaluations show that distance to the Retrained model might not be an adequate evaluation metric for MU.

\section{Data Availability} \label{app:data_availability}
We will provide the code used to perform unlearning given a trained model, compute the metrics for the hyper-parameter searches, and the evaluation metrics.

\section{Requirements} \label{app:requirements}
We ran the experiments on compute clusters with different capacities.
Nonetheless, each method was tested on devices with the same specifications when recording run times: 1 NVIDIA L4 24GB GPU and 4 Intel(R) Xeon(R) CPU @ 2.20GHz.

\section{Visual Summaries}
We  provide a visualization of the Performance Retention as they can take the same values.
\subsection{Performance Retention}
\begin{figure*}
    \centering
    \includegraphics[width=\linewidth]{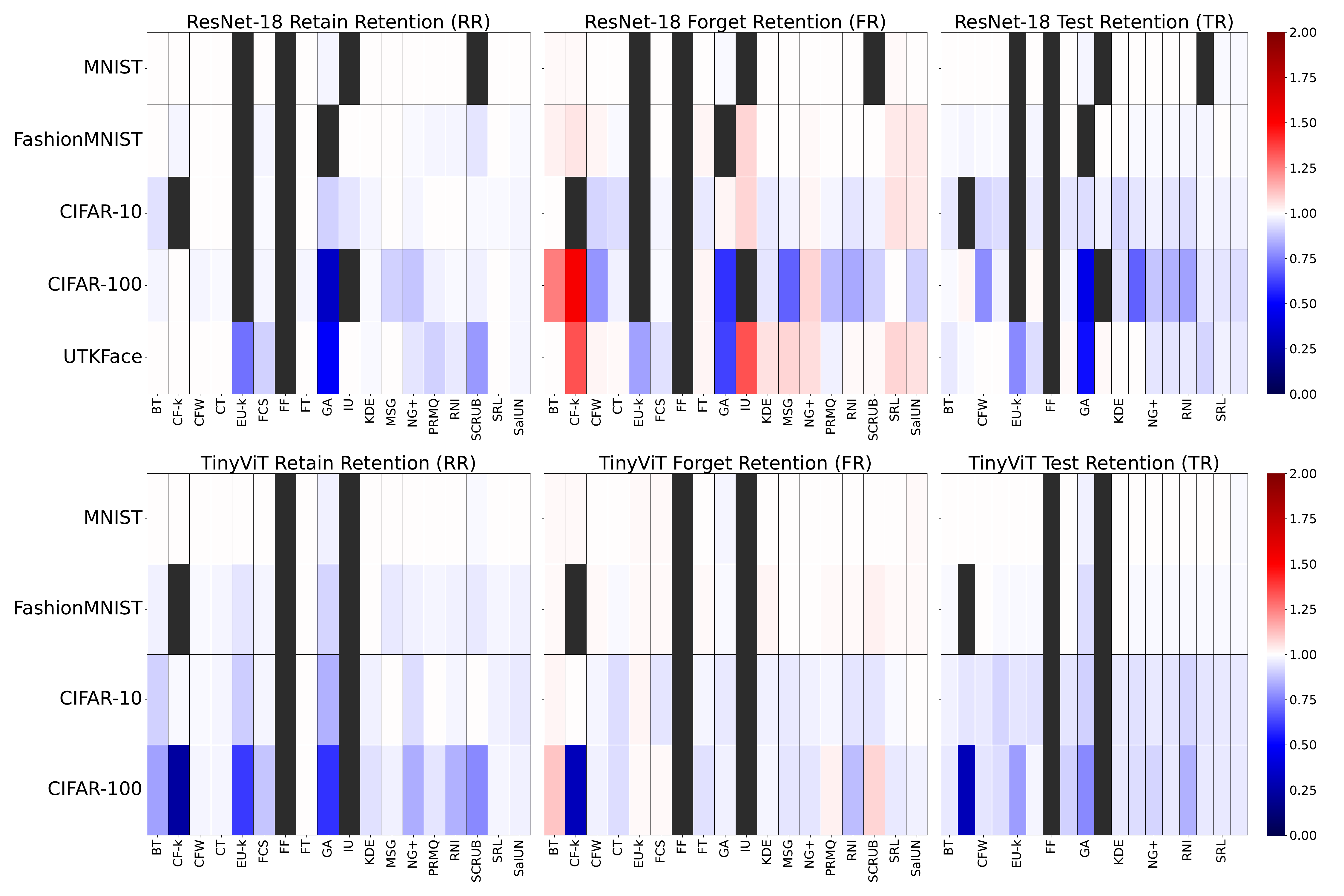}
    \caption{Heatmaps of the performance Retention. (Left) Retain set Retention, (Center) Forget set Retention, (Right) Test Set Retention. (Top) ResNet-18 (Bottom) TinyViT. The optimal value is 1.0, indicating that the model achieves accuracies on par with the reference model.}
    \label{figure:heatmap_retention}
\end{figure*}
\subsection{Radar Plots}
We present Radar Plots illustrating the Retention Deviation using ResNet18 for the methods assessed under U-LIRA.
\begin{figure*}
    \includegraphics[width=\textwidth]{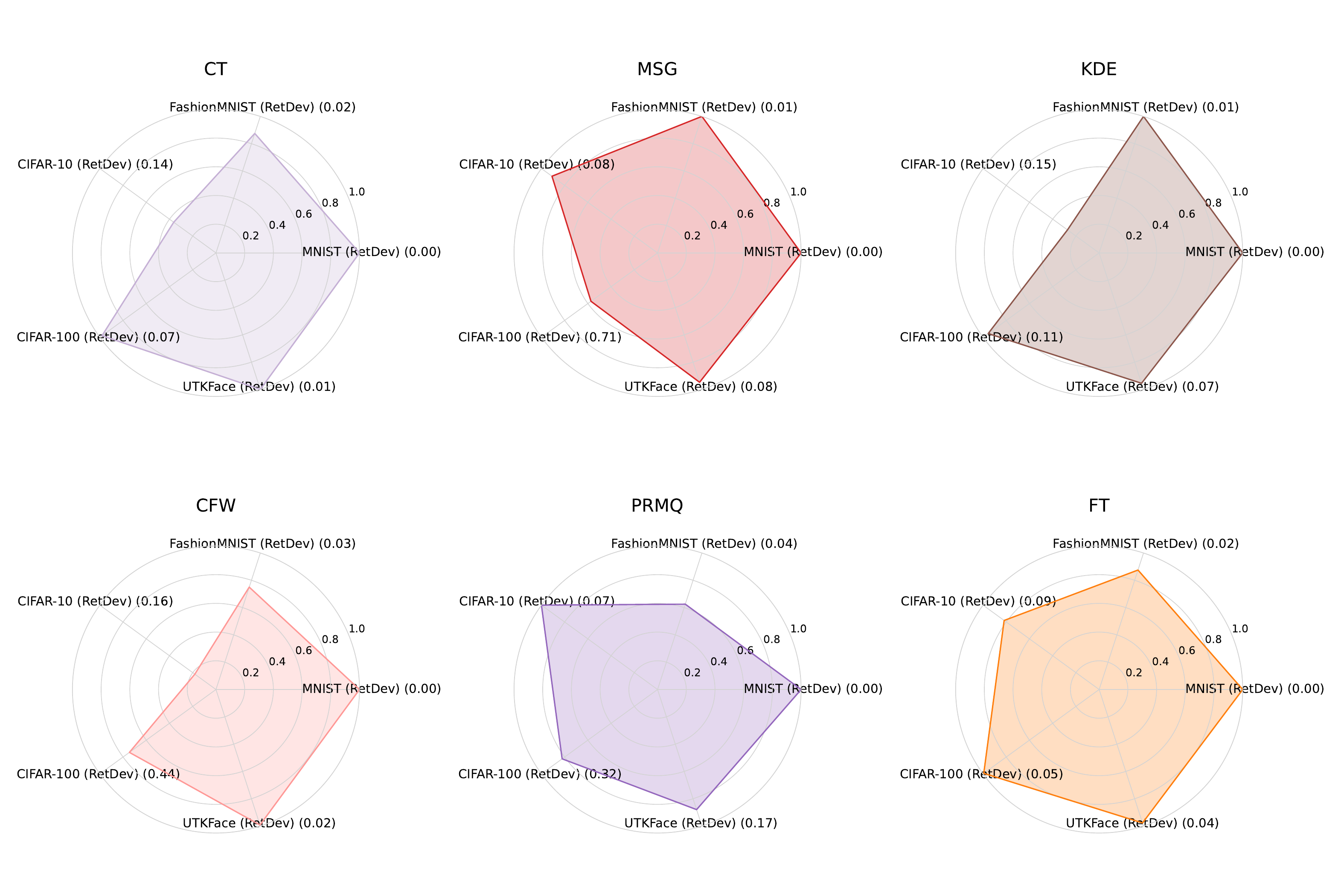}
    \caption{Radar Plot for Retention Deviation across datasets with the ResNet18 architecture (Fuller is better). Since the Retention Deviation takes values in $[0, +\infty)$, we min-max normalize the values, then subtract them from 1 so that lower values lead to fuller plots.}
    \label{figure:radar_retdev}
\end{figure*}

\begin{figure*}
    \includegraphics[width=\textwidth]{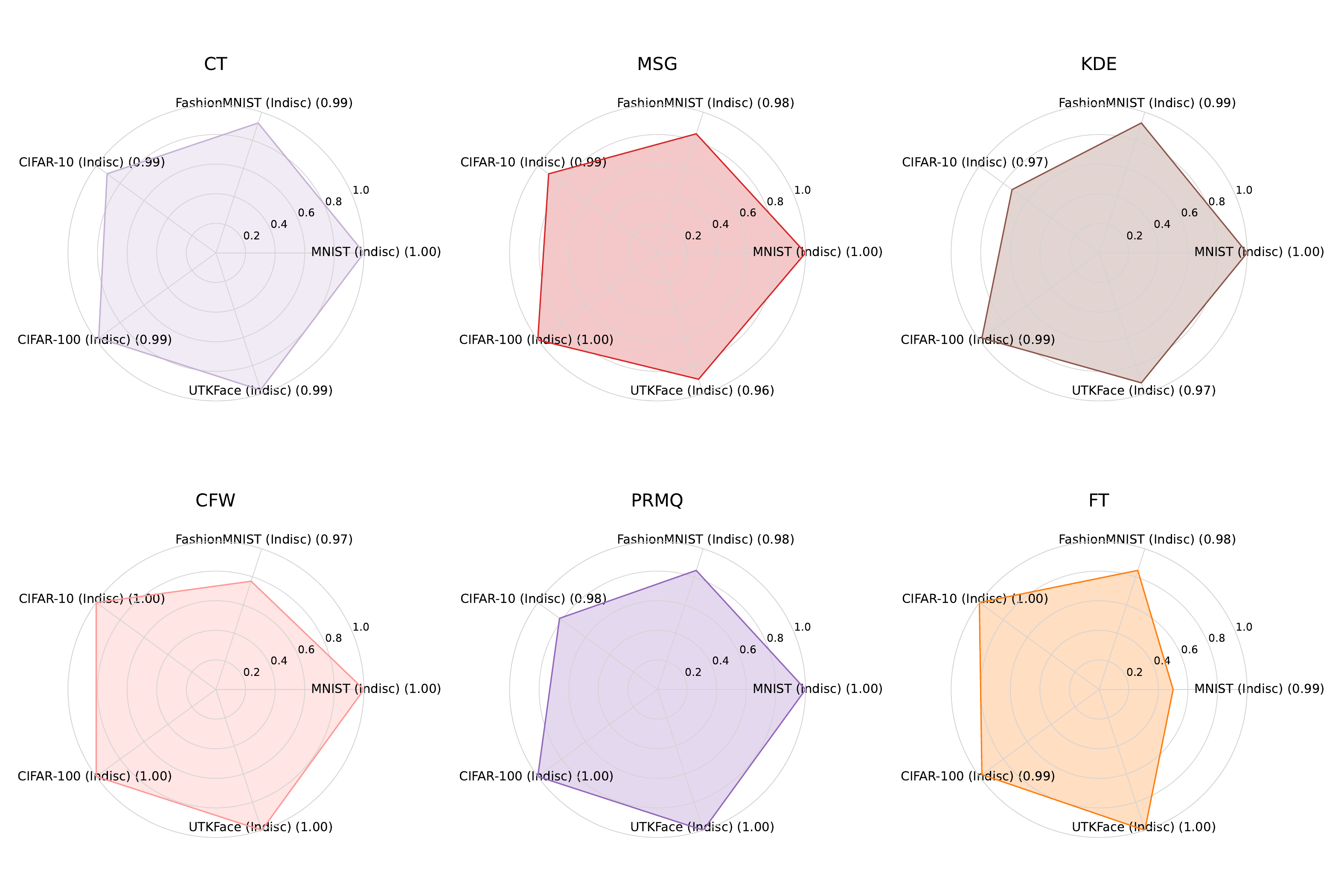}
    \caption{Radar Plot for the Indiscernibility across datasets with the ResNet18 architecture (Fuller is Better). We first apply a min-max normalization across methods.}
    \label{figure:radar_indisc}
\end{figure*}

\end{appendices}

\end{document}